%% file: main.tex
\documentclass{article}
\usepackage{iclr2027_conference,times}
\usepackage[T1]{fontenc}
\usepackage{amsmath,amssymb,booktabs,tabularx,longtable,multirow,graphicx}
\usepackage{tikz,pgfplots,placeins,hyperref,url,etoolbox}
\makeatletter
\let\tableinput\@@input
\makeatother
\usetikzlibrary{arrows.meta,positioning,calc}
\usepgfplotslibrary{groupplots}
\pgfplotsset{compat=1.18}
\input{figures/palette}
\hypersetup{colorlinks=true,linkcolor=blue!50!black,citecolor=blue!50!black,urlcolor=blue!50!black,pdftitle={After the Fix: Transfer of Corrected Agent Experience},pdfauthor={Yanfei Zhang and Xu Lin}}
\iclrfinalcopy
\makeatletter
\patchcmd{\@maketitle}{\lhead{Published as a conference paper at ICLR 2027}}{\lhead{}}{}{\PackageError{arxiv}{Title header patch failed}{}}
\patchcmd{\@maketitle}{\vskip 0.3in minus 0.1in}{\vskip 0in}{}{\PackageError{arxiv}{Title spacing patch failed}{}}
\makeatother
\renewcommand{\headrulewidth}{0pt}

\newcommand{\pp}{\,\text{pp}}
\newlength{\contrastwidth}
\newcolumntype{L}[1]{>{\raggedright\arraybackslash}p{#1}}
\newcolumntype{R}[1]{>{\raggedleft\arraybackslash}p{#1}}
\newcommand{\paperbreak}{\par}
\title{After the Fix: Transfer of Corrected Agent Experience}
\author{Yanfei Zhang\thanks{Equal contribution.}\\
Independent Researcher
\And
Xu Lin\footnotemark[1]\\
International Digital Economy Academy (IDEA)}

\begin{document}
\maketitle
\begin{abstract}
\input{source/abstract_three_regimes}
\end{abstract}

\input{source/visual_intro}
\input{source/visual_related}
\input{source/visual_main}

\clearpage
\section*{Reproducibility Statement}
The analysis uses 3,300 outcome-and-cost records, 1,100 canonical APEX-text reviews, and 20 separate file-state alignment reviews, with paired-analysis scripts, source identifiers, cohort construction, settings, prompts, and editable figures. Every benchmark--condition retains 100 targets. Transition counts and cross-representation consistency are regenerated from the same frozen records.

\section*{Ethics and Data Use}
Feedback is supplied by simulated users; the study involves no human participants. Benchmark materials are used for evaluation, not parameter training. Access and redistribution follow their respective conditions. The statistical package excludes credentials and restricted benchmark attachments.

\section*{AI Use Statement}
Generative AI was used in the experimental pipeline for task pairing, agent execution, simulated user feedback, constructing source-derived experience handoffs, and APEX artifact judging. Model configurations, prompts, and evaluation procedures are documented in Appendices~\ref{app:execution} and~\ref{app:coverage}. OpenAI Codex also assisted with refining research questions, hypotheses, and conceptual framing; implementing and debugging experimental and statistical-analysis code; literature search and synthesis; qualitative trajectory and artifact analysis; interpreting results; and drafting, revising, translating, and formatting the manuscript and figures. Quantitative results are computed by scripts from saved execution and judgment records. Verification includes script-based consistency checks, tests of statistical-analysis routines, and inspection of saved trajectories, artifacts, and cited sources, with judging audits and sensitivity analyses described in Appendix~\ref{app:coverage}. The authors retain responsibility for the research design, interpretation, and final content, including AI-assisted text, code, and artifacts.

\bibliography{references}
\bibliographystyle{iclr2027_conference}
\clearpage
\appendix
\input{appendix}
\end{document}

%% file: figures/palette.tex
\definecolor{paperBlue}{HTML}{3C4DFF}
\definecolor{paperPurple}{HTML}{7642BE}
\definecolor{paperOrange}{HTML}{E2643B}
\definecolor{paperGreen}{HTML}{397C3B}
\definecolor{paperState}{HTML}{94A9D7}
\definecolor{paperBaseline}{HTML}{747A88}

%% file: source/abstract_three_regimes.tex
Does repairing an episode make its experience a better memory for the next task? We transfer the same failed source before and after accepted repair to a fixed target, alongside independent execution. Our 3,300 runs cover 100 ThinkingBox pairs and the same 100 APEX pairs with and without source-state inheritance, under eleven conditions. ThinkingBox's Full/Skill/Hybrid correction gains are 44/29/32 percentage points, with corrected performance 25/22/18 points above independence; inference weakens at the task-family level. Yet 12 of Full's 15-point larger correction gap over Skill come from worse uncorrected performance, not better corrected memory. Moreover, 22 of Full's 46 upward transitions restore observed baseline success. Neither APEX regime establishes comparable aggregate correction benefits. Action evidence connects workflow gains with reusable obligations and convention conflicts with source-local choices. Text APEX's accepted execution reaches 52\% versus its summary's 40\%, without robust global/group-level superiority or an established advantage over independence. Smaller handoffs reduce input but increase calls. The value of repairing experience is therefore distinct from the value of reusing it: memory updates require both a previous-version reference and a fresh-start reference.

%% file: source/visual_intro.tex
\section{Introduction}
An agent repairs a failed task, receives approval, and handles a related request. Its memory now has two versions: the initial failure and an accepted repair. Should the latter replace the former? A system that fixes yesterday's mistake while applying yesterday's assumptions to today's request has improved its source solution without necessarily improving its memory for the next task.

\textbf{The value of repairing experience is not the same as the value of reusing it.} A correction can expose an obligation, such as opening a ticket before acting and closing it afterward, or establish a local workbook version, valuation convention, or input range. The obligation may survive a changed customer; the choice requires renewed grounding. Acceptance validates a source outcome, not the scope of its reuse.

Iterative refinement improves attempts through feedback \citep{madaan2023selfrefine,shinn2023reflexion}; memories and workflows extend reuse across tasks \citep{zhao2024expel,wang2025awm,ouyang2026reasoningbank}. We study \textbf{the incremental transfer value of updating the same failed experience through repair}. This asks whether correction improves a particular memory, beyond whether successful or failed experience can be useful in general.

We study source A at two points: \textbf{U} stops after initial failure; \textbf{C} continues through feedback and repair to acceptance. Both supply the same fixed target B through independent executions. C--U asks whether repair improves the memory; C--B-only asks whether using it beats a fresh start. Pairing the same source holds provenance fixed rather than comparing unrelated successful and failed episodes. A large correction gap can reflect weak U, strong C, or both.

We cross this comparison with how experience is retained. Full history, procedural skill, and hybrid representations select different information; a minimal None condition isolates their shared channel. Summary and Execution ask whether the accepted action segment adds value beyond its outcome summary. Across 3,300 runs, ThinkingBox tests workflow reuse, while matched APEX-state and APEX-text cohorts distinguish implemented handoffs with and without source-artifact inheritance.

\clearpage
\begin{figure}[!t]
\centering
\includegraphics[width=\linewidth]{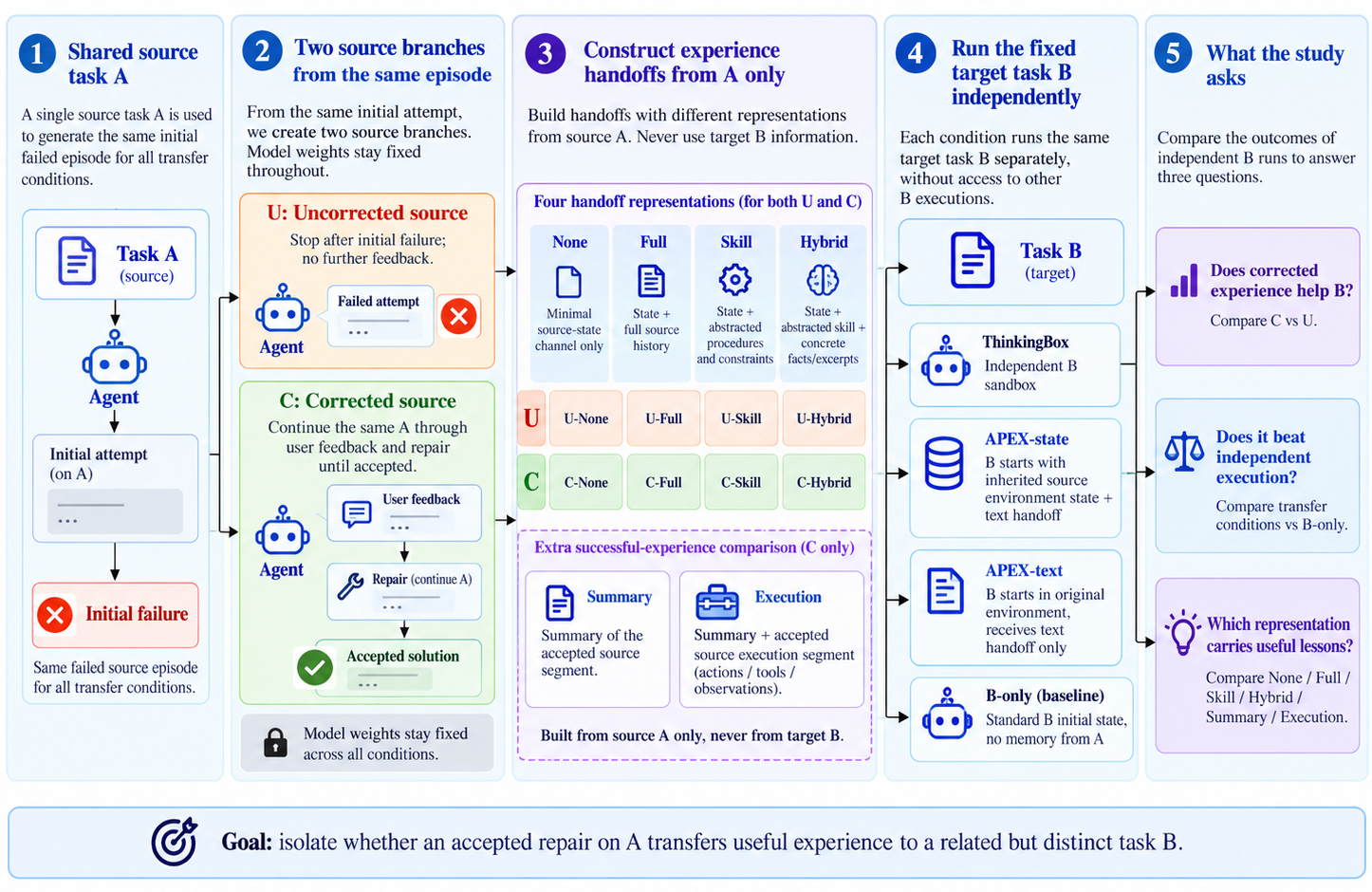}
\caption{The same failed A branches into U/C; each handoff runs B independently. State and text are separate channels. Table~\ref{tab:representation_design} defines all six handoffs, including the benchmark-specific Hybrid constructors. None retains the shared state channel; B-only receives no A experience.}
\label{fig:mechanism}
\end{figure}

Our empirical contribution evaluates \textbf{accepted experience as a memory update}, against both its predecessor and independence. B-only/U/C separates correction gains, baseline restoration, and success beyond both references. Repair-by-representation comparisons show why \textbf{correction sensitivity need not rank corrected-memory quality}: Full's larger gap over Skill mostly reflects weaker U. ThinkingBox's workflow gains and both APEX regimes' weaker aggregate transfer motivate an applicability hypothesis. Tokens and calls distinguish reading savings from execution savings.

%% file: source/visual_related.tex
\section{Related Work}\label{sec:related}
\paragraph{From repairing a task to transferring its correction.}
Self-Refine and Reflexion improve attempts through feedback \citep{madaan2023selfrefine,shinn2023reflexion}; human--AI collaboration distinguishes correction, remembering, and adaptation \citep{amershi2019guidelines}. We compare pre/post-repair experience on a distinct B, making A's acceptance the starting point rather than the transfer outcome.

\paragraph{Experience reuse and representation.}
ExpeL, CLIN, and Agent Workflow Memory extract knowledge and workflows \citep{zhao2024expel,majumder2024clin,wang2025awm}; Trace2Skill studies skill reuse \citep{ni2026trace2skill}. ReasoningBank learns from successes and failures \citep{ouyang2026reasoningbank}. \emph{Break It Down, Pass It On} studies granularity, text/code formats, and negative transfer \citep{feng2026transfer}; SkillsBench measures gains and regressions \citep{li2026skillsbench}. Our axis is \emph{revision of the same failed episode}: fixed-B, pre/post-repair comparisons measure the update's incremental value; B-only separates restoration from success absent under both references.

\paragraph{Content, state, and the work of using memory.}
Prompt compression \citep{jiang2023llmlingua}, masking versus summaries \citep{lindenbauer2025complexity}, repository context \citep{gloaguen2026agents}, and cross-session memory-guided action \citep{he2026memoryarena} connect retention to performance and cost. We compare accepted-segment content, state/text handoffs, and token versus action savings (Appendix~\ref{app:related}).

%% file: source/visual_main.tex
\section{Design: From an Accepted Repair to a New Task}\label{sec:protocol}
Each regime $d$ (ThinkingBox/APEX-state/APEX-text) has $N=100$ source--target pairs $(A_i,B_i)$, indexed by $i=1,\ldots,N$. For branch $s\in\{U,C\}$, $z^s_{Ai}$ records A's history and saved state. The binary source acceptance function $V_{Ai}$ (1: accepted) selects $V_{Ai}(z^U_{Ai})=0$, $V_{Ai}(z^C_{Ai})=1$, with C the first accepted repair within ten feedback--repair rounds.

\paragraph{Two channels of experience.}
For format $r$ (Table~\ref{tab:representation_design}), initialization $\mathcal J$ maps B's original environment $e^0_{Bi}$ and A's record to starting environment $\widetilde e$ and shared state-description text $v$; $\phi$ constructs episode text $m$. Outputs suppress $d$:
\begin{align}
(\widetilde e_{Bi}^{s,r},v_i^{s,r})&=\mathcal J_{d,r}(e^0_{Bi},z^s_{Ai}),\quad m_i^{s,r}=\phi_{d,r}(z^s_{Ai}),\label{eq:handoff}\\
\tau_{Bi}^{s,r}&\sim\operatorname{Rollout}(\pi_\theta,x_{Bi},\widetilde e_{Bi}^{s,r},v_i^{s,r}\oplus m_i^{s,r}).\label{eq:rollout}
\end{align}
Rollout samples ($\sim$) trajectory $\tau$---actions, observations, and outputs---using policy $\pi_\theta$, fixed model parameters $\theta$, B's request $x_{Bi}$, and text concatenation $\oplus$. Text construction sees only A; $v$ can be empty. Only physical inheritance changes $\widetilde e$ from $e^0$.

ThinkingBox supplies terminal-state text to B's independent sandbox. In APEX, $W$ is the original world, $I_A/I_B$ the source/target input bundles, and $\Delta_A^s$ saved source-branch changes. Arrows mean sequential loading/application; $i$ is omitted:
\[
\text{APEX-state}: W\to I_A\to\Delta_A^s\to I_B,
\qquad \text{APEX-text}: W\to I_B.
\]
APEX-state carries 649 filesystem entries and two deletions across 200 A deltas, with no identified database payload. Both regimes share target identities and rubrics.

\subsection{What exactly is handed over?}\label{sec:handoff_contents}
None/Full/Skill/Hybrid use both U and C; Summary/Execution use C alone. With B-only, Table~\ref{tab:representation_design}'s six formats give eleven conditions.
\begin{table}[!htb]\centering\small
\caption{A-derived episode content in each handoff. Every format also retains its regime's shared state channel. No constructor sees B's request or score.}\label{tab:representation_design}
\begin{tabularx}{\linewidth}{@{}lX@{}}\toprule
Format & Episode content given to B\\\midrule
None & Shared state channel only; no additional episode text.\\
Full & Chronological actions, observations, and any failure/feedback/repair, subject to trace limits.\\
Skill & Abstract procedures, constraints, checks, and error--correction rules.\\
Hybrid & ThinkingBox: history summary plus four recent raw assistant segments. APEX: the same Skill plus grounded facts and provenance.\\
Summary & C-only; at most 350 requested words on accepted results, entities, and decisive actions; no earlier failures/feedback.\\
Execution (Exec) & Identical Summary plus the final accepted segment's actions, tool calls, and observations; not the entire repair history.\\\bottomrule
\end{tabularx}\end{table}
None retains terminal-state text in ThinkingBox/APEX-text or restored state in APEX-state; it is not B-only. Text Full/Skill/Hybrid share None's branch prefix. Summary/Execution share initialization and summary, adding only the accepted segment, which may depend on earlier A work. Full--Execution replaces the whole text handoff.

Formats vary in selection, order, abstraction, and length, not a single compression level. U can mark uncertainty and C user validation. C--U changes a correction-associated package; token count and authority cues are not independently manipulated.

\subsection{Questions and contrasts}
\textbf{RQ1: Does repairing A make its experience more useful to B?} H1a/H1b predict that C-Full/Skill/Hybrid outperform their U versions/B-only, respectively. \textbf{RQ2: What should be retained after acceptance?} H2a predicts Execution exceeds Summary; H2b predicts C-Full exceeds Execution. \textbf{RQ3: Does a smaller successful-experience handoff reduce B's work?} Relative to C-Full, H3a/H3b predict that Execution reduces cumulative input tokens/model calls, respectively.

B's binary acceptance function $V_{Bi}$ evaluates its trajectory and outputs: $Y_{di}^{s,r}=V_{Bi}(\tau_{Bi}^{s,r})$; $Y_{di}^{0}$ is B-only acceptance. Mean differences estimate correction (corr: C--U) and baseline-relative (base: C--B-only) gains:
\begin{equation}
\widehat\Delta_{d,r}^{\mathrm{corr}}=\frac1N\sum_{i=1}^{N}(Y_{di}^{C,r}-Y_{di}^{U,r}),\qquad
\widehat\Delta_{d,r}^{\mathrm{base}}=\frac1N\sum_{i=1}^{N}(Y_{di}^{C,r}-Y_{di}^{0}).\label{eq:estimands}
\end{equation}

\section{Evaluation and Statistical Analysis}\label{sec:setup}
ThinkingBox has 65 insurance/35 consulting pairs in 15 domain--family groups. APEX has 48 banking/23 law/29 consulting pairs in 16 worlds: 42 same-family/26 shared-procedure/20 shared-evidence/12 context-only, annotated from full requests. Both APEX regimes share identities and labels: \textbf{200 distinct pairs, not 300 independent pairs} overall.

ThinkingBox uses executable assertions; APEX judges B's answer/artifacts against its original rubric with native visual evidence. Text artifacts reconstruct B's original inputs plus its delta, not A. Each condition retains 100 targets, including budget failures. State/text rubric comparisons align partial-artifact scoring; original judgments and technical provenance remain separately auditable (Appendix~\ref{app:coverage}).

Acceptance uses two-sided exact McNemar tests, with Holm families of nine H1a, nine H1b, and six H2 contrasts, plus global-24 sensitivity. We report raw $p$, family-adjusted $p_H$, and global-adjusted $p_{24}$; pp denotes percentage points. Pointwise paired/group bootstrap intervals retain pairing and 15/16 groups. Exhaustive group sign flips use the same families, requiring independent groups and within-group joint exchangeability under the null, not causal randomization. Exploratory format tests form separate families of 36 within-branch and 18 correction-gap contrasts. Rubric means weight tasks equally. Analyses are retrospective, not preregistered.

APEX uses \texttt{deepseek-flash} with thinking, 100 steps, and 7,200 seconds; ThinkingBox defaults to \texttt{deepseek-v4-flash}, high reasoning, 400 interactions, and 1,200 seconds. Regime-specific settings are fixed (Appendix~\ref{app:execution}).

\paperbreak
\section{RQ1: When Does Corrected Experience Transfer?}\label{sec:results}
\begin{table}[!htb]\centering\small
\caption{Absolute target acceptance (\%). Every cell has 100 targets, including budget failures. State/Text are the two APEX regimes.}\label{tab:absolute_main}
\begin{tabular*}{\linewidth}{@{\extracolsep{\fill}}lrrr@{}}\toprule
Condition & ThinkingBox & State & Text\\\midrule
B-only & 42 & 44 & 47\\
U-None / C-None & 37 / 36 & 45 / 44 & 45 / 47\\
U-Full / C-Full & 23 / 67 & 42 / 41 & 45 / 45\\
U-Skill / C-Skill & 35 / 64 & 39 / 42 & 44 / 45\\
U-Hybrid / C-Hybrid & 28 / 60 & 41 / 45 & 48 / 40\\
Summary / Execution & 56 / 67 & 40 / 44 & 40 / 52\\\bottomrule
\end{tabular*}\end{table}
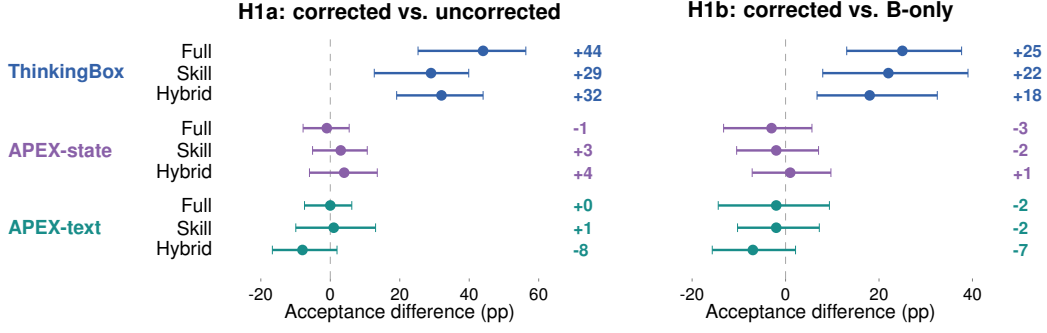
\begin{figure}[!htb]\centering
\resizebox{\linewidth}{!}{\input{figures/rq1_forest_en}}
\caption{Dots/bars: paired acceptance differences/pointwise 95\% cluster-bootstrap intervals; 100 targets/row. Numeric labels are estimates, not adjusted significance.}\label{fig:correction_map}
\end{figure}
\subsection{Correction helps across representations in ThinkingBox, but not generally in APEX}
\textbf{An accepted repair is useful in one setting without being a general transfer signal.} ThinkingBox C-Full accepts 67 of 100 targets, compared with 23 for U-Full and 42 for B-only. Repair therefore improves on both the failed memory and independent execution, rather than merely making a bad handoff less bad. Skill and Hybrid show the same ordering: their corrected versions accept 64 and 60 targets, above both their U versions (35 and 28) and B-only. In contrast, APEX-state C-Full accepts 41 targets against U-Full's 42 and B-only's 44; APEX-text C-Full and U-Full both accept 45, below B-only's 47. Figure~\ref{fig:correction_map} shows that neither APEX regime establishes a comparable aggregate advantage across the three representations. Source acceptance alone thus does not distinguish an update worth reusing from one that only solved A.

\textbf{The ThinkingBox pattern is not confined to one representation or one family.} For 42 targets, changing U to C turns failure into success under at least two of Full, Skill, and Hybrid; 18 improve under all three. The corresponding counts are only five and one in APEX-state, and six and one in APEX-text (Table~\ref{tab:consistency}). This asks whether the same targets benefit when the repaired episode is represented differently, not merely whether three aggregate means rise. ThinkingBox's overlap points to useful content surviving different representations. Its C--U advantages also remain positive after removing any single family, so no one family alone accounts for the observed direction. 

\paperbreak
\subsection{A correction gain can restore a success that memory displaced}
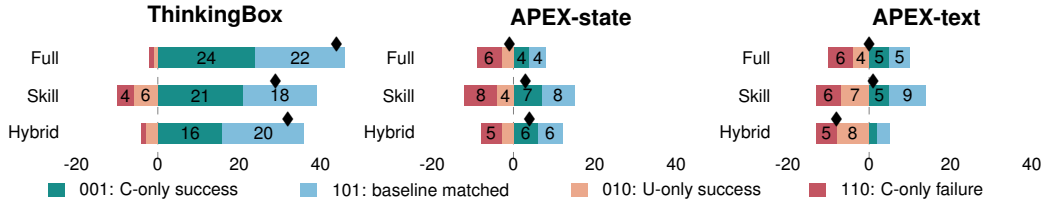
\begin{figure}[!htb]\centering
\resizebox{\linewidth}{!}{\input{figures/rq1_decomposition_en}}
\caption{B-only/U/C transitions per 100 targets, not individual causal effects. Bars distinguish baseline outcomes; diamonds show net C--U. Table~\ref{tab:rq1paths} includes unchanged patterns.}\label{fig:decomposition_main}
\end{figure}
Fix $d,r$ and omit them below. Let $n_{buc}$ count targets with B-only/U/C outcomes $(b,u,c)\in\{0,1\}^3$, where 1 denotes acceptance. Net transfer decomposes as
\[
N\widehat\Delta^{\mathrm{corr}}=
\underbrace{n_{001}}_{\text{C-only success}}+
\underbrace{n_{101}}_{\text{baseline matched}}-
\underbrace{n_{010}}_{\text{U-only success}}-
\underbrace{n_{110}}_{\text{C-only failure}}.
\]
ThinkingBox Full produces 46 upward transitions from U to C and two downward transitions, giving its $44\pp$ net gain. But 22 of those 46 improvements occur where B-only already succeeds: U loses that success and C restores it. The other 24 occur where both B-only and U fail. These are different reasons to value correction. Restoration justifies replacing a misleading memory already in use; success beyond both references supports introducing the corrected experience instead of starting independently. Reporting only C--U merges these decisions and would count nearly half of Full's upward transitions as if they were successes unavailable without memory.

\textbf{Conversely, a zero mean need not mean that correction changes nothing.} APEX-text Full has ten U-failure/C-success transitions and ten U-success/C-failure transitions: the acceptance rate stays at 45\%, while twenty targets change outcome. APEX-state Full similarly has eight gains and nine losses. There is room to improve---text U-Full fails on 55 targets---but the successes recovered by C are offset by successes lost elsewhere. Thus the APEX result is not simply saturation or complete irrelevance of the handoff. It is an unfavorable balance between helping some targets and preserving performance on others. The transition counts are observed paired outcomes, not individual causal effects, but they identify what an aggregate transfer score conceals.

\paperbreak
\subsection{The representation that improves most is not necessarily the best to reuse}\label{sec:representations}
\textbf{Full's larger correction gain mostly reflects its weaker starting point.} In ThinkingBox, Full moves from 23\% under U to 67\% under C; Skill moves from 35\% to 64\%. Full's correction gain is therefore fifteen points larger, although its corrected acceptance is only three points higher. Twelve of the fifteen points arise because U-Full performs worse than U-Skill. The three-point C-Full advantage is not established by the adjusted representation tests (Appendix~\ref{app:statistics}). This separates two questions: how strongly a representation responds to repair, and which corrected representation should be supplied to B. Ranking the first quantity as though it answered the second would overstate the case for retaining the full history.

\textbf{None shows that episode content matters; it does not tell us how much detail to keep.} ThinkingBox C-None retains the terminal-state description and accepts 36 targets. C-Full, C-Skill, and C-Hybrid accept 67, 64, and 60. Their shared advantage over this minimal channel means the observed benefit is not reproduced by reporting the source's terminal state alone. Yet the three richer formats need not preserve the same information or solve the same targets. APEX-text makes this distinction concrete: C-Full and C-Skill each accept 45 targets, but disagree on 24, with twelve successes exclusive to each. Equal acceptance therefore does not make compression behaviorally neutral; the abstract representation exchanges some successes for others rather than simply retaining a smaller equivalent memory.

Hybrid tests whether restoring detail to an abstraction improves that tradeoff. APEX Hybrid adds grounded facts to the same Skill, but corrected acceptance moves from 42\% to 45\% with state inheritance and from 45\% to 40\% without it. Neither adjusted difference is established. The results offer no general ordering in which adding facts is always better, just as they do not establish Skill's noninferiority to Full. Full retains context and competing attempts; Skill foregrounds procedures; Hybrid restores selected particulars. Their practical difference is which evidence B can use to decide whether a source lesson applies, not merely how many tokens are removed. The behavioral cases below examine that decision directly.

\paperbreak
\subsection{Behavior distinguishes reusable obligations from source-specific choices}\label{sec:behavior}
\begin{figure}[!htb]\centering
\resizebox{\linewidth}{!}{\input{figures/rq1_case_chain_en}}
\caption{Selected action chains, not frequencies: lifecycle (ThinkingBox), discounting/precision (APEX). Both APEX regimes show the latter; displayed decisions come from text trajectories.}\label{fig:case_chain_main}
\end{figure}
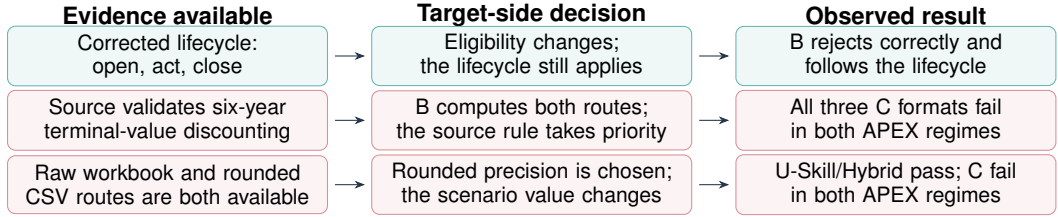
\textbf{A workflow obligation can remain useful even when the answer changes.} ThinkingBox's lifecycle rule requires opening a ticket, acting, and then closing it. In \texttt{lif\_017}$\to$\texttt{lif\_015}, B's eligibility differs from A's, so copying A's business decision is inappropriate. B-only and U-Skill already reject B correctly, but create an already-solved ticket; C-Skill rejects correctly and follows the required lifecycle. Correction contributes how to carry out the decision, not which eligibility answer to copy. The broader validator pattern is consistent with this example: lifecycle violations fall from 55 to two under Full, 40 to seven under Skill, and 48 to three under Hybrid. Of the 54 flags removed under Full, 37 targets pass and 17 still fail. A reusable procedure can therefore explain an important behavioral improvement without accounting for every remaining task requirement.

\textbf{An accepted local choice can instead override the new request.} In APEX pair \texttt{ad52b8019140}, B explicitly requests mid-year discounting. Text C-Skill describes full-year terminal discounting as user-approved. B computes both the six-year and 5.5-year alternatives, yet selects the former, producing 40.45/44.83 rather than 41.62/46.22. The missing capability is not calculating the alternative: it appears in the trace. The failure is selecting a source-approved convention despite the target's different requirement. B-only and all U formats pass in both APEX regimes, while all C formats receive zero credit. Its presence with and without inherited artifacts shows that this pattern is not confined to reconstructing missing source files.

The contrast is about \emph{scope}, not a rule that detailed memories are harmful. In Helios, correction supplies the useful scope restriction: text U-Hybrid combines 4,900 SKU rows, whereas C-Hybrid uses B's specified 2,500-row Rebuilt input and recovers the correct result. Both APEX regimes show B-only pass, U-Hybrid fail, C-Hybrid pass, making this a concrete instance of baseline restoration. Correction can thus narrow attention to the right input or elevate the wrong convention. Source approval alone does not tell B which of these it is doing; the target requirement must resolve the choice.

These traces motivate preserving an obligation together with the conditions that justify it. They do not establish a causal effect of an approval marker or a population frequency of this mechanism. Precision and workbook-lineage cases also require care: Lumea's rounded-input choice changes the result, whereas 3M's target does not specify which workbook version to use. A lower fixed-rubric score in the latter is not sufficient evidence of harmful reuse. Appendix~\ref{app:cases} retains these selected, non-blind audits; Appendix~\ref{app:relationships} reports domain and state/text comparisons. The main distinction is supported by target-specific action evidence: \emph{carrying over a procedure that still applies is different from carrying over a decision whose premises have changed}.

\paperbreak
\section{RQ2 and RQ3: Action Detail, Acceptance, and Target-Side Cost}\label{sec:boundary}\label{sec:cost}
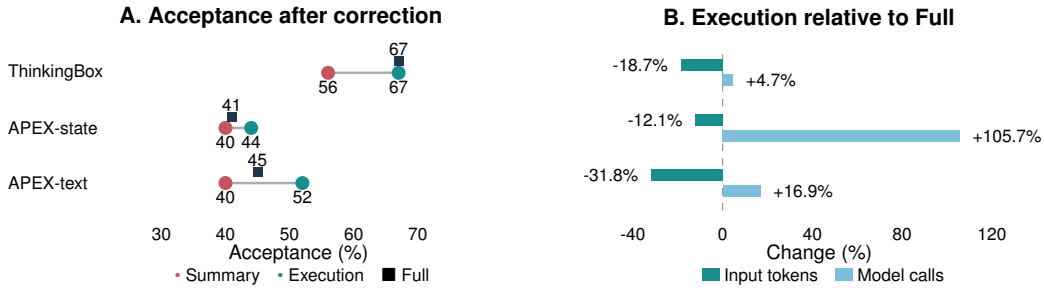
\begin{figure}[!htb]\centering
\resizebox{\linewidth}{!}{\input{figures/retention_cost_en}}
\caption{Acceptance per 100 targets (left); mean Full-to-Execution input/call changes (right). Tables~\ref{tab:regimeboundary}--\ref{tab:regimecost} report exact means and tests.}\label{fig:retention_cost_main}
\end{figure}
\subsection{Action detail improves on a summary, not necessarily on starting fresh}
Summary and Execution share the same accepted-source summary and initialization; Execution additionally supplies the accepted segment's actions, tool calls, and observations. Their contrast asks whether showing how A was completed adds value beyond describing its result. In APEX-text, Execution accepts 52 targets against Summary's 40: fourteen targets change from failure to success and two change in the opposite direction. ThinkingBox likewise rises from 56 to 67, whereas APEX-state rises from 40 to 44 (Figure~\ref{fig:retention_cost_main}). The common positive direction favors retaining operational evidence, but its magnitude and statistical support differ across settings.

Text Execution's twelve-point advantage over Summary is only five points over B-only's 47\%, with a world interval crossing zero for the latter. Improving a summary-based handoff therefore does not establish superiority over independent execution. Text H2a survives target-level H2-family adjustment ($p_H=0.0251$), but not global-24 or group adjustment; the other regimes do not pass H2-family adjustment. Full also establishes no advantage over Execution. The clearest observed benefit is thus adding action evidence to a summary, not retaining the entire repair history.

\subsection{Naming a requirement is different from carrying it out}
Two ThinkingBox gains make the added evidence interpretable. Summary already instructs B to set the ticket type to \texttt{task}, yet B fails to set the field; with Execution, it does. The requirement was available in both handoffs, so these cases are not explained by its absence from the summary. Execution additionally shows concrete operations and observations. The contrast suggests that a handoff can fail between recognizing a requirement and implementing it, making operational examples useful even when the summary names the right rule. It does not isolate which part of the added segment produced the improvement.

\subsection{A smaller handoff reduces context reading but can require more execution}
Relative to Full, Execution reads less cumulative input in all three regimes, yet makes more model calls. APEX-text illustrates the tradeoff: mean input falls from 3.576M to 2.438M tokens (31.8\%), while calls rise from 30.01 to 35.08 (16.9\%) and tool calls from 38.24 to 44.40. The explanation at the accounting level is that input per call falls from 119.2k to 69.5k: each interaction reads less context, enough to outweigh the additional interactions. H3a and H3b therefore separate. Reducing repeated context consumption is not the same achievement as reducing the work needed to complete B.

APEX-state approximately doubles calls under Execution (20.04 to 41.22), limiting its input reduction to 12.1\%; ThinkingBox's calls rise only 4.7\% while input falls 18.7\%. More work remains downstream, although extra calls need not all be reconstruction. Which representation is preferable depends on the constraint: in text APEX, 32 accepted Execution runs versus eleven Full runs fit within two million input tokens, but only ten versus eighteen fit within twenty model calls. These are completed-run counts, not early-stopping experiments. Execution also uses 12.2\% more input than Summary. Memory selection must therefore consider success, context consumption, and interaction demand jointly. Appendix~\ref{app:retention} gives uncertainty and accounting boundaries; B-stage token reductions are not billing or whole-pipeline savings.

\paperbreak
\FloatBarrier
\section{Discussion: What Does Repair Add to Reusable Experience?}\label{sec:discussion}
\paragraph{The same-episode comparison evaluates a revision, not a category of memories.}
Learning from successful and failed experience is already central to agent memory \citep{ouyang2026reasoningbank,feng2026transfer}. Our comparison asks what is gained by repairing \emph{a particular failed episode}. Holding its provenance and target fixed reveals a distinction that source acceptance alone cannot resolve: a revision may improve on its predecessor without being preferable to independent execution. C--U addresses the first decision; C--B-only addresses the second. A system selecting memory updates needs both.

ThinkingBox Full moves from 23\% to 67\%, whereas Skill moves from 35\% to 64\%. Their 44- and 29-point correction gains would make Full look substantially better if only C--U were considered. Yet twelve of the fifteen extra points come from Full's worse uncorrected version; the corrected versions differ by only three points. A larger response to repair can reflect a weaker predecessor, rather than a better final representation.

\paragraph{Repair can recover performance that memory itself displaced.}
Full's 46 upward transitions contain two different outcomes. On 22 targets, B-only already succeeds, U fails, and C succeeds: repair recovers the result available without source experience. On the other 24, both B-only and U fail before C succeeds. Both count as gains over U, but only the second pattern records success beyond both references. Restoration remains valuable when misleading memory is already in use; it is not the same justification for introducing memory in the first place. For memory updates, the useful question is consequently not just how many failures were fixed, but which previous successes were retained and which successes became available beyond a fresh start.

\paragraph{A repaired solution mixes general obligations with local decisions.}
Behavior traces suggest why source improvement and transfer value diverge. In ThinkingBox, a changed eligibility decision still requires the same open--act--close workflow: the obligation remains applicable when the answer changes. In APEX, the source-approved full-year convention conflicts with B's explicit mid-year requirement. B computes both alternatives but selects the source convention. The problem is not a missing procedure; it is using a locally valid choice outside the conditions that justified it. Both APEX regimes show this case, while neither establishes ThinkingBox-like aggregate correction gains.

This distinction suggests a concrete memory-design objective: preserve the useful correction without turning every accepted source decision into a default rule. An entry would retain what was changed, the requirement that justified the change, and the inputs or artifacts on which it depended. The target agent would then check whether those conditions still hold. For example, a ticket lifecycle can remain relevant across customers, whereas a workbook version or discounting convention must be checked against the new request. Acceptance supplies evidence about what worked on A; B's requirements determine what can be reused.

\paragraph{Representation and state change the work required to make that decision.}
Full can supply context for reconstructing a rule's scope; Skill makes the rule easier to find but can omit that context. Hybrid adds detail, which helps only if it supports the right target-side decision. Execution can show how to carry out an action that Summary merely names. The choice is not simply more versus less memory \citep{lindenbauer2025complexity}. APEX-state makes source files available, whereas APEX-text can require rebuilding them; neither establishes that their assumptions fit B. In Planet Fitness, reconstruction changes performance without changing C--U. A useful handoff must support both selecting and applying appropriate knowledge. Its cost includes the resulting target work: our smaller Execution handoffs reduce input but increase calls.

\section{Scope and Open Questions}\label{sec:open_questions}
The next question is whether explicitly representing applicability improves transfer. A focused test would compare the same C-Skill rules unchanged, with a generic reminder to check B, and with source-derived conditions specifying when each rule applies, holding the model and budget fixed. The reminder separates the value of those conditions from simply encouraging more caution. On held-out targets, with constructors restricted to A, evaluation should measure whether convention conflicts decrease \emph{while useful workflow behavior is retained}. Generic-checklist and approval-marker controls would distinguish shared norms, task-specific lessons, and source authority.

The present evidence concerns complete repair packages, failed-but-repairable sources, and one execution per cell. Group-level inference addresses task dependence, not repeat-run reliability; repetitions and blinded artifact adjudication would distinguish persistent gains from execution and scoring variation. Crossing state with identical text would isolate inheritance, and actual database changes would extend the file-state evidence. A longer-term question is whether successive repairs accumulate useful knowledge or obsolete commitments. Studying target-aware reuse with natural human feedback, including supervision, memory construction, and reconstruction costs, would connect the current one-handoff comparison to a continuing memory system.

\section{Conclusion}
Repairing a failed episode and obtaining useful experience from it are distinct achievements. The same-episode design shows why: a large correction gain may reflect a poor uncorrected memory, and an improved target outcome may restore performance already available without memory. ThinkingBox also shows success beyond both references, while the two APEX regimes do not establish comparable aggregate benefits. Repair matters, but source acceptance alone does not say when its lessons should be reused.

The resulting principle is to evaluate a memory revision against both its predecessor and a fresh start, and to preserve the conditions under which its advice was valid. This shifts the design question from storing an accepted solution to deciding which parts should guide the next task. After the fix, useful experience must help B choose what still applies, not merely reproduce what worked on A.
\label{maintextend}

%% file: figures/rq1_forest_en.tex
\begin{tikzpicture}[x=1cm,y=0.9cm,font=\sffamily\fontsize{8}{9}\selectfont]
\definecolor{ink}{HTML}{243447}
\definecolor{newgain}{HTML}{188D89}
\definecolor{restore}{HTML}{80BDDC}
\definecolor{losthelp}{HTML}{EBA88C}
\definecolor{broken}{HTML}{C35562}
\node[font=\sffamily\bfseries\small] at (5.825,4.630) {H1a: corrected vs. uncorrected};
\draw[gray!70,dashed] (4.814,0.310) -- (4.814,4.230);
\draw[gray] (3.803,0.260) -- (3.803,0.310);
\node[font=\sffamily\scriptsize] at (3.803,0.060) {-20};
\draw[gray] (4.814,0.260) -- (4.814,0.310);
\node[font=\sffamily\scriptsize] at (4.814,0.060) {0};
\draw[gray] (5.825,0.260) -- (5.825,0.310);
\node[font=\sffamily\scriptsize] at (5.825,0.060) {20};
\draw[gray] (6.836,0.260) -- (6.836,0.310);
\node[font=\sffamily\scriptsize] at (6.836,0.060) {40};
\draw[gray] (7.847,0.260) -- (7.847,0.310);
\node[font=\sffamily\scriptsize] at (7.847,0.060) {60};
\node[] at (5.825,-0.240) {Acceptance difference (pp)};
\definecolor{cohort}{HTML}{3361A8}
\node[anchor=west,text=cohort,font=\sffamily\bfseries\fontsize{8}{9}\selectfont] at (0.000,3.630) {ThinkingBox};
\node[anchor=east] at (3.220,3.990) {Full};
\draw[cohort,line width=1pt] (6.091,3.990) -- (7.660,3.990);
\draw[cohort] (6.091,3.920) -- (6.091,4.060);
\draw[cohort] (7.660,3.920) -- (7.660,4.060);
\fill[cohort] (7.038,3.990) circle (2.2pt);
\node[anchor=west,text=cohort,font=\sffamily\bfseries\scriptsize] at (8.230,3.990) {+44};
\node[anchor=east] at (3.220,3.630) {Skill};
\draw[cohort,line width=1pt] (5.454,3.630) -- (6.829,3.630);
\draw[cohort] (5.454,3.560) -- (5.454,3.700);
\draw[cohort] (6.829,3.560) -- (6.829,3.700);
\fill[cohort] (6.280,3.630) circle (2.2pt);
\node[anchor=west,text=cohort,font=\sffamily\bfseries\scriptsize] at (8.230,3.630) {+29};
\node[anchor=east] at (3.220,3.270) {Hybrid};
\draw[cohort,line width=1pt] (5.780,3.270) -- (7.038,3.270);
\draw[cohort] (5.780,3.200) -- (5.780,3.340);
\draw[cohort] (7.038,3.200) -- (7.038,3.340);
\fill[cohort] (6.432,3.270) circle (2.2pt);
\node[anchor=west,text=cohort,font=\sffamily\bfseries\scriptsize] at (8.230,3.270) {+32};
\definecolor{cohort}{HTML}{8960A8}
\node[anchor=west,text=cohort,font=\sffamily\bfseries\fontsize{8}{9}\selectfont] at (0.000,2.380) {APEX-state};
\node[anchor=east] at (3.220,2.740) {Full};
\draw[cohort,line width=1pt] (4.418,2.740) -- (5.089,2.740);
\draw[cohort] (4.418,2.670) -- (4.418,2.810);
\draw[cohort] (5.089,2.670) -- (5.089,2.810);
\fill[cohort] (4.763,2.740) circle (2.2pt);
\node[anchor=west,text=cohort,font=\sffamily\bfseries\scriptsize] at (8.230,2.740) {-1};
\node[anchor=east] at (3.220,2.380) {Skill};
\draw[cohort,line width=1pt] (4.556,2.380) -- (5.352,2.380);
\draw[cohort] (4.556,2.310) -- (4.556,2.450);
\draw[cohort] (5.352,2.310) -- (5.352,2.450);
\fill[cohort] (4.966,2.380) circle (2.2pt);
\node[anchor=west,text=cohort,font=\sffamily\bfseries\scriptsize] at (8.230,2.380) {+3};
\node[anchor=east] at (3.220,2.020) {Hybrid};
\draw[cohort,line width=1pt] (4.511,2.020) -- (5.498,2.020);
\draw[cohort] (4.511,1.950) -- (4.511,2.090);
\draw[cohort] (5.498,1.950) -- (5.498,2.090);
\fill[cohort] (5.016,2.020) circle (2.2pt);
\node[anchor=west,text=cohort,font=\sffamily\bfseries\scriptsize] at (8.230,2.020) {+4};
\definecolor{cohort}{HTML}{188D89}
\node[anchor=west,text=cohort,font=\sffamily\bfseries\fontsize{8}{9}\selectfont] at (0.000,1.130) {APEX-text};
\node[anchor=east] at (3.220,1.490) {Full};
\draw[cohort,line width=1pt] (4.439,1.490) -- (5.127,1.490);
\draw[cohort] (4.439,1.420) -- (4.439,1.560);
\draw[cohort] (5.127,1.420) -- (5.127,1.560);
\fill[cohort] (4.814,1.490) circle (2.2pt);
\node[anchor=west,text=cohort,font=\sffamily\bfseries\scriptsize] at (8.230,1.490) {+0};
\node[anchor=east] at (3.220,1.130) {Skill};
\draw[cohort,line width=1pt] (4.312,1.130) -- (5.473,1.130);
\draw[cohort] (4.312,1.060) -- (4.312,1.200);
\draw[cohort] (5.473,1.060) -- (5.473,1.200);
\fill[cohort] (4.864,1.130) circle (2.2pt);
\node[anchor=west,text=cohort,font=\sffamily\bfseries\scriptsize] at (8.230,1.130) {+1};
\node[anchor=east] at (3.220,0.770) {Hybrid};
\draw[cohort,line width=1pt] (3.971,0.770) -- (4.912,0.770);
\draw[cohort] (3.971,0.700) -- (3.971,0.840);
\draw[cohort] (4.912,0.700) -- (4.912,0.840);
\fill[cohort] (4.409,0.770) circle (2.2pt);
\node[anchor=west,text=cohort,font=\sffamily\bfseries\scriptsize] at (8.230,0.770) {-8};
\node[font=\sffamily\bfseries\small] at (11.950,4.630) {H1b: corrected vs. B-only};
\draw[gray!70,dashed] (11.440,0.310) -- (11.440,4.230);
\draw[gray] (10.080,0.260) -- (10.080,0.310);
\node[font=\sffamily\scriptsize] at (10.080,0.060) {-20};
\draw[gray] (11.440,0.260) -- (11.440,0.310);
\node[font=\sffamily\scriptsize] at (11.440,0.060) {0};
\draw[gray] (12.800,0.260) -- (12.800,0.310);
\node[font=\sffamily\scriptsize] at (12.800,0.060) {20};
\draw[gray] (14.160,0.260) -- (14.160,0.310);
\node[font=\sffamily\scriptsize] at (14.160,0.060) {40};
\node[] at (11.950,-0.240) {Acceptance difference (pp)};
\definecolor{cohort}{HTML}{3361A8}
\draw[cohort,line width=1pt] (12.330,3.990) -- (14.002,3.990);
\draw[cohort] (12.330,3.920) -- (12.330,4.060);
\draw[cohort] (14.002,3.920) -- (14.002,4.060);
\fill[cohort] (13.140,3.990) circle (2.2pt);
\node[anchor=west,text=cohort,font=\sffamily\bfseries\scriptsize] at (14.630,3.990) {+25};
\draw[cohort,line width=1pt] (11.981,3.630) -- (14.095,3.630);
\draw[cohort] (11.981,3.560) -- (11.981,3.700);
\draw[cohort] (14.095,3.560) -- (14.095,3.700);
\fill[cohort] (12.936,3.630) circle (2.2pt);
\node[anchor=west,text=cohort,font=\sffamily\bfseries\scriptsize] at (14.630,3.630) {+22};
\draw[cohort,line width=1pt] (11.898,3.270) -- (13.648,3.270);
\draw[cohort] (11.898,3.200) -- (11.898,3.340);
\draw[cohort] (13.648,3.200) -- (13.648,3.340);
\fill[cohort] (12.664,3.270) circle (2.2pt);
\node[anchor=west,text=cohort,font=\sffamily\bfseries\scriptsize] at (14.630,3.270) {+18};
\definecolor{cohort}{HTML}{8960A8}
\draw[cohort,line width=1pt] (10.538,2.740) -- (11.824,2.740);
\draw[cohort] (10.538,2.670) -- (10.538,2.810);
\draw[cohort] (11.824,2.670) -- (11.824,2.810);
\fill[cohort] (11.236,2.740) circle (2.2pt);
\node[anchor=west,text=cohort,font=\sffamily\bfseries\scriptsize] at (14.630,2.740) {-3};
\draw[cohort,line width=1pt] (10.728,2.380) -- (11.920,2.380);
\draw[cohort] (10.728,2.310) -- (10.728,2.450);
\draw[cohort] (11.920,2.310) -- (11.920,2.450);
\fill[cohort] (11.304,2.380) circle (2.2pt);
\node[anchor=west,text=cohort,font=\sffamily\bfseries\scriptsize] at (14.630,2.380) {-2};
\draw[cohort,line width=1pt] (10.954,2.020) -- (12.100,2.020);
\draw[cohort] (10.954,1.950) -- (10.954,2.090);
\draw[cohort] (12.100,1.950) -- (12.100,2.090);
\fill[cohort] (11.508,2.020) circle (2.2pt);
\node[anchor=west,text=cohort,font=\sffamily\bfseries\scriptsize] at (14.630,2.020) {+1};
\definecolor{cohort}{HTML}{188D89}
\draw[cohort,line width=1pt] (10.460,1.490) -- (12.078,1.490);
\draw[cohort] (10.460,1.420) -- (10.460,1.560);
\draw[cohort] (12.078,1.420) -- (12.078,1.560);
\fill[cohort] (11.304,1.490) circle (2.2pt);
\node[anchor=west,text=cohort,font=\sffamily\bfseries\scriptsize] at (14.630,1.490) {-2};
\draw[cohort,line width=1pt] (10.741,1.130) -- (11.931,1.130);
\draw[cohort] (10.741,1.060) -- (10.741,1.200);
\draw[cohort] (11.931,1.060) -- (11.931,1.200);
\fill[cohort] (11.304,1.130) circle (2.2pt);
\node[anchor=west,text=cohort,font=\sffamily\bfseries\scriptsize] at (14.630,1.130) {-2};
\draw[cohort,line width=1pt] (10.373,0.770) -- (11.585,0.770);
\draw[cohort] (10.373,0.700) -- (10.373,0.840);
\draw[cohort] (11.585,0.700) -- (11.585,0.840);
\fill[cohort] (10.964,0.770) circle (2.2pt);
\node[anchor=west,text=cohort,font=\sffamily\bfseries\scriptsize] at (14.630,0.770) {-7};
\end{tikzpicture}

%% file: figures/rq1_decomposition_en.tex
\begin{tikzpicture}[x=1cm,y=0.9cm,font=\sffamily\fontsize{8}{9}\selectfont]
\definecolor{ink}{HTML}{243447}
\definecolor{newgain}{HTML}{188D89}
\definecolor{restore}{HTML}{80BDDC}
\definecolor{losthelp}{HTML}{EBA88C}
\definecolor{broken}{HTML}{C35562}
\node[font=\sffamily\bfseries\small] at (2.650,2.580) {ThinkingBox};
\draw[gray,dashed] (1.843,0.350) -- (1.843,2.200);
\node[anchor=east,font=\sffamily\scriptsize] at (0.580,1.940) {Full};
\fill[newgain] (1.843,1.780) rectangle (3.214,2.100);
\node[font=\sffamily\scriptsize] at (2.529,1.940) {24};
\fill[restore] (3.214,1.780) rectangle (4.471,2.100);
\node[font=\sffamily\scriptsize] at (3.843,1.940) {22};
\fill[losthelp] (1.786,1.780) rectangle (1.843,2.100);
\fill[broken] (1.729,1.780) rectangle (1.786,2.100);
\node[font=\scriptsize] at (4.357,2.160) {$\blacklozenge$};
\node[anchor=east,font=\sffamily\scriptsize] at (0.580,1.350) {Skill};
\fill[newgain] (1.843,1.190) rectangle (3.043,1.510);
\node[font=\sffamily\scriptsize] at (2.443,1.350) {21};
\fill[restore] (3.043,1.190) rectangle (4.071,1.510);
\node[font=\sffamily\scriptsize] at (3.557,1.350) {18};
\fill[losthelp] (1.500,1.190) rectangle (1.843,1.510);
\node[font=\sffamily\scriptsize] at (1.671,1.350) {6};
\fill[broken] (1.271,1.190) rectangle (1.500,1.510);
\node[font=\sffamily\scriptsize] at (1.386,1.350) {4};
\node[font=\scriptsize] at (3.500,1.570) {$\blacklozenge$};
\node[anchor=east,font=\sffamily\scriptsize] at (0.580,0.760) {Hybrid};
\fill[newgain] (1.843,0.600) rectangle (2.757,0.920);
\node[font=\sffamily\scriptsize] at (2.300,0.760) {16};
\fill[restore] (2.757,0.600) rectangle (3.900,0.920);
\node[font=\sffamily\scriptsize] at (3.329,0.760) {20};
\fill[losthelp] (1.671,0.600) rectangle (1.843,0.920);
\fill[broken] (1.614,0.600) rectangle (1.671,0.920);
\node[font=\scriptsize] at (3.671,0.980) {$\blacklozenge$};
\node[font=\sffamily\scriptsize] at (0.700,0.280) {-20};
\node[font=\sffamily\scriptsize] at (1.843,0.280) {0};
\node[font=\sffamily\scriptsize] at (2.986,0.280) {20};
\node[font=\sffamily\scriptsize] at (4.129,0.280) {40};
\node[font=\sffamily\bfseries\small] at (7.650,2.580) {APEX-state};
\draw[gray,dashed] (6.843,0.350) -- (6.843,2.200);
\node[anchor=east,font=\sffamily\scriptsize] at (5.580,1.940) {Full};
\fill[newgain] (6.843,1.780) rectangle (7.071,2.100);
\node[font=\sffamily\scriptsize] at (6.957,1.940) {4};
\fill[restore] (7.071,1.780) rectangle (7.300,2.100);
\node[font=\sffamily\scriptsize] at (7.186,1.940) {4};
\fill[losthelp] (6.671,1.780) rectangle (6.843,2.100);
\fill[broken] (6.329,1.780) rectangle (6.671,2.100);
\node[font=\sffamily\scriptsize] at (6.500,1.940) {6};
\node[font=\scriptsize] at (6.786,2.160) {$\blacklozenge$};
\node[anchor=east,font=\sffamily\scriptsize] at (5.580,1.350) {Skill};
\fill[newgain] (6.843,1.190) rectangle (7.243,1.510);
\node[font=\sffamily\scriptsize] at (7.043,1.350) {7};
\fill[restore] (7.243,1.190) rectangle (7.700,1.510);
\node[font=\sffamily\scriptsize] at (7.471,1.350) {8};
\fill[losthelp] (6.614,1.190) rectangle (6.843,1.510);
\node[font=\sffamily\scriptsize] at (6.729,1.350) {4};
\fill[broken] (6.157,1.190) rectangle (6.614,1.510);
\node[font=\sffamily\scriptsize] at (6.386,1.350) {8};
\node[font=\scriptsize] at (7.014,1.570) {$\blacklozenge$};
\node[anchor=east,font=\sffamily\scriptsize] at (5.580,0.760) {Hybrid};
\fill[newgain] (6.843,0.600) rectangle (7.186,0.920);
\node[font=\sffamily\scriptsize] at (7.014,0.760) {6};
\fill[restore] (7.186,0.600) rectangle (7.529,0.920);
\node[font=\sffamily\scriptsize] at (7.357,0.760) {6};
\fill[losthelp] (6.671,0.600) rectangle (6.843,0.920);
\fill[broken] (6.386,0.600) rectangle (6.671,0.920);
\node[font=\sffamily\scriptsize] at (6.529,0.760) {5};
\node[font=\scriptsize] at (7.071,0.980) {$\blacklozenge$};
\node[font=\sffamily\scriptsize] at (5.700,0.280) {-20};
\node[font=\sffamily\scriptsize] at (6.843,0.280) {0};
\node[font=\sffamily\scriptsize] at (7.986,0.280) {20};
\node[font=\sffamily\scriptsize] at (9.129,0.280) {40};
\node[font=\sffamily\bfseries\small] at (12.650,2.580) {APEX-text};
\draw[gray,dashed] (11.843,0.350) -- (11.843,2.200);
\node[anchor=east,font=\sffamily\scriptsize] at (10.580,1.940) {Full};
\fill[newgain] (11.843,1.780) rectangle (12.129,2.100);
\node[font=\sffamily\scriptsize] at (11.986,1.940) {5};
\fill[restore] (12.129,1.780) rectangle (12.414,2.100);
\node[font=\sffamily\scriptsize] at (12.271,1.940) {5};
\fill[losthelp] (11.614,1.780) rectangle (11.843,2.100);
\node[font=\sffamily\scriptsize] at (11.729,1.940) {4};
\fill[broken] (11.271,1.780) rectangle (11.614,2.100);
\node[font=\sffamily\scriptsize] at (11.443,1.940) {6};
\node[font=\scriptsize] at (11.843,2.160) {$\blacklozenge$};
\node[anchor=east,font=\sffamily\scriptsize] at (10.580,1.350) {Skill};
\fill[newgain] (11.843,1.190) rectangle (12.129,1.510);
\node[font=\sffamily\scriptsize] at (11.986,1.350) {5};
\fill[restore] (12.129,1.190) rectangle (12.643,1.510);
\node[font=\sffamily\scriptsize] at (12.386,1.350) {9};
\fill[losthelp] (11.443,1.190) rectangle (11.843,1.510);
\node[font=\sffamily\scriptsize] at (11.643,1.350) {7};
\fill[broken] (11.100,1.190) rectangle (11.443,1.510);
\node[font=\sffamily\scriptsize] at (11.271,1.350) {6};
\node[font=\scriptsize] at (11.900,1.570) {$\blacklozenge$};
\node[anchor=east,font=\sffamily\scriptsize] at (10.580,0.760) {Hybrid};
\fill[newgain] (11.843,0.600) rectangle (11.957,0.920);
\fill[restore] (11.957,0.600) rectangle (12.129,0.920);
\fill[losthelp] (11.386,0.600) rectangle (11.843,0.920);
\node[font=\sffamily\scriptsize] at (11.614,0.760) {8};
\fill[broken] (11.100,0.600) rectangle (11.386,0.920);
\node[font=\sffamily\scriptsize] at (11.243,0.760) {5};
\node[font=\scriptsize] at (11.386,0.980) {$\blacklozenge$};
\node[font=\sffamily\scriptsize] at (10.700,0.280) {-20};
\node[font=\sffamily\scriptsize] at (11.843,0.280) {0};
\node[font=\sffamily\scriptsize] at (12.986,0.280) {20};
\node[font=\sffamily\scriptsize] at (14.129,0.280) {40};
\fill[newgain] (0.30,-.23) rectangle (0.53,-.02);
\node[anchor=west,font=\sffamily\scriptsize] at (0.620,-0.130) {001: C-only success};
\fill[restore] (3.85,-.23) rectangle (4.08,-.02);
\node[anchor=west,font=\sffamily\scriptsize] at (4.170,-0.130) {101: baseline matched};
\fill[losthelp] (7.65,-.23) rectangle (7.88,-.02);
\node[anchor=west,font=\sffamily\scriptsize] at (7.970,-0.130) {010: U-only success};
\fill[broken] (11.00,-.23) rectangle (11.23,-.02);
\node[anchor=west,font=\sffamily\scriptsize] at (11.320,-0.130) {110: C-only failure};
\end{tikzpicture}

%% file: figures/rq1_case_chain_en.tex
\begin{tikzpicture}[x=1cm,y=0.8cm,font=\sffamily\fontsize{8}{9}\selectfont]
\definecolor{ink}{HTML}{243447}
\definecolor{newgain}{HTML}{188D89}
\definecolor{restore}{HTML}{80BDDC}
\definecolor{losthelp}{HTML}{EBA88C}
\definecolor{broken}{HTML}{C35562}
\node[font=\sffamily\bfseries\small] at (2.200,3.650) {Evidence available};
\node[font=\sffamily\bfseries\small] at (7.050,3.650) {Target-side decision};
\node[font=\sffamily\bfseries\small] at (11.900,3.650) {Observed result};
\node[draw=newgain!65,fill=newgain!7,rounded corners=3pt,align=center,text width=4.05cm,minimum height=.77cm,inner sep=3pt] at (2.200,2.970) {Corrected lifecycle:\\open, act, close};
\draw[-{Stealth[length=4pt]},ink] (4.430,2.970) -- (4.830,2.970);
\node[draw=newgain!65,fill=newgain!7,rounded corners=3pt,align=center,text width=4.05cm,minimum height=.77cm,inner sep=3pt] at (7.050,2.970) {Eligibility changes;\\the lifecycle still applies};
\draw[-{Stealth[length=4pt]},ink] (9.280,2.970) -- (9.680,2.970);
\node[draw=newgain!65,fill=newgain!7,rounded corners=3pt,align=center,text width=4.05cm,minimum height=.77cm,inner sep=3pt] at (11.900,2.970) {B rejects correctly and\\follows the lifecycle};
\node[draw=broken!65,fill=broken!7,rounded corners=3pt,align=center,text width=4.05cm,minimum height=.77cm,inner sep=3pt] at (2.200,1.900) {Source validates six-year\\terminal-value discounting};
\draw[-{Stealth[length=4pt]},ink] (4.430,1.900) -- (4.830,1.900);
\node[draw=broken!65,fill=broken!7,rounded corners=3pt,align=center,text width=4.05cm,minimum height=.77cm,inner sep=3pt] at (7.050,1.900) {B computes both routes;\\the source rule takes priority};
\draw[-{Stealth[length=4pt]},ink] (9.280,1.900) -- (9.680,1.900);
\node[draw=broken!65,fill=broken!7,rounded corners=3pt,align=center,text width=4.05cm,minimum height=.77cm,inner sep=3pt] at (11.900,1.900) {All three C formats fail\\in both APEX regimes};
\node[draw=broken!65,fill=broken!7,rounded corners=3pt,align=center,text width=4.05cm,minimum height=.77cm,inner sep=3pt] at (2.200,0.830) {Raw workbook and rounded\\CSV routes are both available};
\draw[-{Stealth[length=4pt]},ink] (4.430,0.830) -- (4.830,0.830);
\node[draw=broken!65,fill=broken!7,rounded corners=3pt,align=center,text width=4.05cm,minimum height=.77cm,inner sep=3pt] at (7.050,0.830) {Rounded precision is chosen;\\the scenario value changes};
\draw[-{Stealth[length=4pt]},ink] (9.280,0.830) -- (9.680,0.830);
\node[draw=broken!65,fill=broken!7,rounded corners=3pt,align=center,text width=4.05cm,minimum height=.77cm,inner sep=3pt] at (11.900,0.830) {U-Skill/Hybrid pass; C fail\\in both APEX regimes};
\end{tikzpicture}

%% file: figures/retention_cost_en.tex
\begin{tikzpicture}[x=1cm,y=0.9cm,font=\sffamily\fontsize{8}{9}\selectfont]
\definecolor{ink}{HTML}{243447}
\definecolor{newgain}{HTML}{188D89}
\definecolor{restore}{HTML}{80BDDC}
\definecolor{losthelp}{HTML}{EBA88C}
\definecolor{broken}{HTML}{C35562}
\node[font=\sffamily\bfseries\small] at (4.000,3.480) {A. Acceptance after correction};
\node[font=\sffamily\bfseries\small] at (11.500,3.480) {B. Execution relative to Full};
\draw[gray,dashed] (10.275,0.430) -- (10.275,2.900);
\node[anchor=west,font=\sffamily\scriptsize] at (0.000,2.600) {ThinkingBox};
\draw[gray!65,line width=1pt] (4.669,2.600) -- (5.671,2.600);
\node[circle,fill=broken,inner sep=2pt] at (4.669,2.600) {};
\node[font=\sffamily\scriptsize] at (4.669,2.380) {56};
\node[circle,fill=newgain,inner sep=2pt] at (5.671,2.600) {};
\node[font=\sffamily\scriptsize] at (5.671,2.380) {67};
\node[rectangle,fill=ink,inner sep=2pt] at (5.671,2.780) {};
\node[font=\sffamily\scriptsize] at (5.671,2.970) {67};
\fill[newgain] (10.275,2.630) rectangle (9.679,2.810);
\node[anchor=east,font=\sffamily\scriptsize] at (9.619,2.720) {-18.7\%};
\fill[restore] (10.275,2.390) rectangle (10.426,2.570);
\node[anchor=west,font=\sffamily\scriptsize] at (10.486,2.480) {+4.7\%};
\node[anchor=west,font=\sffamily\scriptsize] at (0.000,1.730) {APEX-state};
\draw[gray!65,line width=1pt] (3.211,1.730) -- (3.576,1.730);
\node[circle,fill=broken,inner sep=2pt] at (3.211,1.730) {};
\node[font=\sffamily\scriptsize] at (3.211,1.510) {40};
\node[circle,fill=newgain,inner sep=2pt] at (3.576,1.730) {};
\node[font=\sffamily\scriptsize] at (3.576,1.510) {44};
\node[rectangle,fill=ink,inner sep=2pt] at (3.302,1.910) {};
\node[font=\sffamily\scriptsize] at (3.302,2.100) {41};
\fill[newgain] (10.275,1.760) rectangle (9.889,1.940);
\node[anchor=east,font=\sffamily\scriptsize] at (9.829,1.850) {-12.1\%};
\fill[restore] (10.275,1.520) rectangle (13.644,1.700);
\node[anchor=west,font=\sffamily\scriptsize] at (13.704,1.610) {+105.7\%};
\node[anchor=west,font=\sffamily\scriptsize] at (0.000,0.860) {APEX-text};
\draw[gray!65,line width=1pt] (3.211,0.860) -- (4.304,0.860);
\node[circle,fill=broken,inner sep=2pt] at (3.211,0.860) {};
\node[font=\sffamily\scriptsize] at (3.211,0.640) {40};
\node[circle,fill=newgain,inner sep=2pt] at (4.304,0.860) {};
\node[font=\sffamily\scriptsize] at (4.304,0.640) {52};
\node[rectangle,fill=ink,inner sep=2pt] at (3.667,1.040) {};
\node[font=\sffamily\scriptsize] at (3.667,1.230) {45};
\fill[newgain] (10.275,0.890) rectangle (9.260,1.070);
\node[anchor=east,font=\sffamily\scriptsize] at (9.200,0.980) {-31.8\%};
\fill[restore] (10.275,0.650) rectangle (10.814,0.830);
\node[anchor=west,font=\sffamily\scriptsize] at (10.874,0.740) {+16.9\%};
\node[font=\sffamily\scriptsize] at (2.300,0.060) {30};
\node[font=\sffamily\scriptsize] at (3.211,0.060) {40};
\node[font=\sffamily\scriptsize] at (4.122,0.060) {50};
\node[font=\sffamily\scriptsize] at (5.033,0.060) {60};
\node[font=\sffamily\scriptsize] at (5.944,0.060) {70};
\node[font=\sffamily\scriptsize] at (9.000,0.060) {-40};
\node[font=\sffamily\scriptsize] at (10.275,0.060) {0};
\node[font=\sffamily\scriptsize] at (11.550,0.060) {40};
\node[font=\sffamily\scriptsize] at (12.825,0.060) {80};
\node[font=\sffamily\scriptsize] at (14.100,0.060) {120};
\node[] at (4.250,-0.260) {Acceptance (\%)};
\node[] at (11.650,-0.260) {Change (\%)};
\node[font=\sffamily\scriptsize] at (4.300,-0.620) {\textcolor{broken}{\textbullet} Summary\quad\textcolor{newgain}{\textbullet} Execution\quad$\blacksquare$ Full};
\node[font=\sffamily\scriptsize] at (11.700,-0.620) {\textcolor{newgain}{\rule{7pt}{5pt}} Input tokens\quad\textcolor{restore}{\rule{7pt}{5pt}} Model calls};
\end{tikzpicture}

%% file: appendix.tex
\input{source/appendix_protocol}
\FloatBarrier
\input{source/appendix_statistics}
\FloatBarrier
\input{source/appendix_rq1}
\FloatBarrier
\input{source/appendix_cases}
\FloatBarrier
\input{source/appendix_retention}
\FloatBarrier
\input{source/appendix_judging}
\FloatBarrier
\input{source/appendix_related}
\FloatBarrier

%% file: source/appendix_protocol.tex
\section{Protocol, Cohorts, and Execution Settings}\label{app:execution}
Equations~\ref{eq:handoff}--\ref{eq:rollout} define the initialization and textual channels. Table~\ref{tab:components} specifies their implementation in each regime, and Table~\ref{tab:boundarydesign} distinguishes the quantities held fixed by each contrast. Source records and constructors are regime-specific; memory construction has no access to B's request.

\begin{table}[!htb]\centering\small
\caption{State and text channels. $W$ is the original APEX world, $I_A/I_B$ the task inputs, and $\Delta_A^s$ branch-specific source changes. B-only excludes A in every regime.}\label{tab:components}\label{tab:settings}
\begin{tabularx}{\linewidth}{@{}lXX@{}}\toprule
Regime & B's starting environment & A-derived information\\\midrule
ThinkingBox & Independent standard B sandbox & Terminal-state description and the specified textual representation; no live A database restoration\\
APEX-state & $W\rightarrow I_A\rightarrow\Delta_A^s\rightarrow I_B$ & Source environment state plus the specified text; None supplies the state channel alone\\
APEX-text & $W\rightarrow I_B$ & Text alone; None supplies terminal observations, with the same branch prefix in Full/Skill/Hybrid\\\bottomrule
\end{tabularx}\end{table}
\FloatBarrier

\begin{table}[!htb]\centering\small
\caption{Experimental contrasts and their interpretation.}\label{tab:boundarydesign}
\begin{tabularx}{\linewidth}{@{}lXX@{}}\toprule
Contrast & Held fixed & What the contrast measures\\\midrule
C--U, same representation & Target and evaluation & The complete correction-associated experience package\\
C--B-only & Target and evaluation & Benefit beyond independent target execution\\
Execution--Summary & Exact summary, framing, starting state & Adding the accepted segment, including its content and length\\
Full--Execution & Target and evaluation & Alternative whole handoffs, not history alone\\
APEX-text--APEX-state & Pair identities and target rubric & Implemented handoff regimes, with state and textual-construction differences\\\bottomrule
\end{tabularx}\end{table}
\FloatBarrier
\paragraph{ThinkingBox selection.}
The source benchmark is ThinkingBox \citep{li2026thinkingbox}.
Task identifiers are grouped by domain and workflow-family prefix, shuffled with seed 260914, and paired cyclically within each family before formal A outcomes are observed. Screening includes 79 insurance and 81 consulting candidates: 19/46 initially pass, leaving 60/35 failures. Five pilot failures supplement these 95. Eight sources remain unaccepted after ten feedback rounds and are replaced by eight unused natural pilot failures. The final cohort contains 87 formal-screen and 13 pilot sources, not a held-out cohort. Its accepted repairs occur at rounds one through four for 78/13/6/3 sources. The full task mapping is included in \path{data/cohort_construction.json}.

ThinkingBox retries restore A's canonical sandbox because erroneous writes to append-only audit records cannot always be undone; visible failure, feedback, and repair are retained in the episode. Each B starts in its own standard sandbox. This differs from physically carrying A's database forward and is why the state channel is textual in Equations~\ref{eq:handoff}--\ref{eq:rollout}. Feedback privately uses the executed trace and validator differences, asks for the earliest causal mistake, and must translate it into user-facing language without exposing reference actions, assertions, or scores. It is simulated feedback with privileged diagnostic access, not a study of unaided human review.

\paragraph{Shared APEX cohort.}
Both regimes use the APEX-Agents benchmark \citep{vidgen2026apex}.
The pairing stage assigns a within-world one-to-one mapping with no self-pairs across 480 tasks in 33 worlds. The pairer sees requests and delivery modes, not outcomes, gold answers, or rubric scores; malformed mappings are completed by constrained assignment using suggested pairs and lexical similarity. These prospective pairings are distinct from the retrospective relationship annotations in Appendix~\ref{app:relationships}. Five pilot A attempts are reused. Of the other 475, 462 complete and are reviewed: 227 initially pass and 235 are correction candidates. The selection gate records 198 accepted repairs plus three accepted pilot repairs, giving 201 eligible pairs. Both APEX regimes use the same first 100 pairs in frozen manifest order; selection is not optimized on B results and yields 16 worlds. All C branches use their first saved accepted repair. The cohort estimates utility conditional on failure, repairability, and selection order, not population performance across all APEX-state tasks.

\paragraph{Source feedback and compression.}
APEX-state review inspects the task, rubric, hidden reference when available, final answer, tool trace, and changed artifacts. It produces criterion judgments and natural-language corrective feedback; hidden reference values are not intentionally passed through the feedback prompt. Memory generation receives the source request, visible user turns, final answer, tool digest, and changed artifacts, not B. Its prompt distinguishes user-validated corrections from uncertain unresolved tactics. Skill stores procedural knowledge; Hybrid appends concrete facts and provenance to the same Skill. ThinkingBox compression produces a state-and-action summary (up to 220 words), procedural skill (450 words), and chronological summary (900 words); Hybrid combines the latter with four recent assistant-turn segments. Full follows per-message and total trace truncation. These are requested budgets, not guaranteed token counts.

\begin{table}[!htb]
\centering\footnotesize
\caption{Execution settings. Output budgets are tokens per call. The package contains full prompt constructors and saved configuration fields.}
\begin{tabularx}{\linewidth}{@{}lXX@{}}\toprule
Component & ThinkingBox & APEX-state\\\midrule
Acting model & Environment-resolved DeepSeek alias; adapter default \texttt{deepseek-v4-flash} & DeepSeek V4.1 Flash (multimodal), API \texttt{deepseek-flash}\\
Agent sampling & Thinking enabled; explicit \texttt{reasoning\_effort=high}; temperature omitted & Thinking enabled; no explicit reasoning effort; temperature 0.2; no supplied seed\\
Agent budgets & 32,768 output; 400 interactions; 30 user turns; 1,200 seconds & 8,192 output; 100 steps; 7,200 seconds\\
Runtime user & Temperature 0.2; 4,096 output; cannot terminate agent conversation & No separate conversational user during B\\
Correction review & Temperature 0.2; 2,048 output; trace and state differences & Temperature 0; 8,192 output; artifacts and criteria\\
Memory compressor & Temperature 0; 8,192 output; thinking disabled & \texttt{deepseek-v4-pro}; temperature 0; 8,192 output\\
B acceptance & Executable task assertions; configured legacy judge where applicable & Owner review; delivery integrity and all criteria satisfied\\
Version record & Resolved alias and provider revision unavailable & All 1,100 target summaries record \texttt{deepseek-flash}\\\bottomrule
\end{tabularx}\label{tab:execution}
\end{table}
The prompts are exported from the implementation: \texttt{REVIEWER\_SYSTEM}, the U/C memory constructors, terminal-state extraction, APEX-state \texttt{SYSTEM\_PROMPT}, review, compression, and pairing, plus the runtime user's instructions. The retained ThinkingBox terminal-field rerun container records \texttt{MODEL\_NAME=deepseek-v4-flash}; earlier deleted containers' resolved settings cannot be inferred from it. The package excludes credentials and benchmark attachments. Re-execution requires authorized materials, sandbox services, and runner dependencies. An immutable historical API model cannot be reconstructed from an unrecorded revision.

\paragraph{Successful-experience handoffs.}
The Success-summary constructors for both benchmarks are target-blind and request at most 350 words covering observable results, entities, and decisive actions. Earlier failures and user corrections are excluded from this summary input. Execution adds the final accepted source segment to the identical summary. In ThinkingBox the accepted attempt starts from A's restored standard sandbox; in APEX-state it can be a final repair over artifacts already modified by preceding attempts. Thus an accepted segment is not necessarily an end-to-end replay from the original environment. APEX-state retains up to 16,000 characters per message and 140,000 characters overall. Both added APEX-state conditions restore identical C-state changes before B; ThinkingBox uses B's standard sandbox. Full reuses C-Full's original prompt, not a nested extension of Execution.

\paragraph{Text-only initialization and state inventory.}
APEX-text starts B from the original world and B inputs, supplying A-derived text without restoring A's inputs or source delta. Its acting model is \texttt{deepseek-flash}, with thinking enabled, 100 steps, 8,192 output tokens, and a 7,200-second task timeout. It uses the same 100 pair identities and target rubrics as APEX-state. Terminal observations and Full serialization differ across implementations, so the comparison is between implemented handoff regimes, not an isolated state intervention.

The inspected APEX-state payloads contain 649 filesystem files and two deletions across 200 unique A deltas; A inputs contain 45 filesystem files and one \texttt{.DS\_Store}. No \texttt{.apps\_data}, SQLite header, or database filename is observed. The realized transfer is therefore file-state transfer; support for database restoration in the environment does not establish that database content was transferred in this cohort.

%% file: source/appendix_statistics.tex
\section{Complete Results and Statistical Comparisons}\label{app:threetables}\label{app:statistics}
Every regime--condition has 100 targets. ThinkingBox measures executable acceptance; APEX reports owner acceptance and the mean fraction of rubric criteria satisfied per task. APEX-state uses the partial-artifact alignment layer described in Appendix~\ref{app:coverage}; unfinished executions retain their original status and the full denominator.
\begin{table}[!htb]\centering\small
\caption{All eleven conditions in all three regimes. Each entry has denominator 100. A/Q denotes acceptance percentage / task-normalized rubric percentage.}\label{tab:regimeall}
\begin{tabular*}{\linewidth}{@{\extracolsep{\fill}}lrrr@{}}\toprule
Condition & ThinkingBox & APEX-state A/Q & APEX-text A/Q\\\midrule
\tableinput data/regime_pairing/all_rows.tex
\bottomrule\end{tabular*}\end{table}

\subsection{Paired estimands and comparison families}
For candidate and reference outcomes on the same target, $G$ counts reference failures becoming successes and $L$ the reverse. The acceptance difference is $(G-L)/N$, while $(G+L)/N$ measures outcome reversal. Exact two-sided McNemar tests use the discordant pairs. Paired bootstrap intervals resample targets together with all their conditions; cluster intervals resample the 15 ThinkingBox domain--family groups or 16 APEX worlds. The archived ThinkingBox/APEX-state intervals use 20,000 resamples; text and matched-regime analyses use 10,000. All are pointwise, not simultaneous intervals.

The retrospective manuscript-wide Holm families contain nine H1a contrasts (C--U), nine H1b contrasts (C--B-only), and six H2 contrasts (Execution--Summary and Full--Execution). Table~\ref{tab:integratedcontrasts} also reports global Holm over all 24 tests. All six ThinkingBox H1 contrasts survive the global adjustment. APEX-text Execution--Summary has family $p=0.0251$ and global $p=0.0753$: its significance depends on the comparison scope. Domain and relationship analyses are exploratory.

\begingroup\footnotesize\setlength{\tabcolsep}{4pt}
\setlength{\contrastwidth}{\dimexpr\linewidth-14\tabcolsep\relax}
\begin{longtable}{@{}L{.085\contrastwidth}L{.175\contrastwidth}R{.08\contrastwidth}R{.08\contrastwidth}L{.20\contrastwidth}R{.125\contrastwidth}R{.125\contrastwidth}R{.13\contrastwidth}@{}}
\caption{All 24 paired acceptance contrasts. G/L counts gains/losses; differences and world/family-cluster 95\% intervals are percentage points. TB/State/Text denote the three regimes; F/S/H in condition names denote Full/Skill/Hybrid, and standalone B/S/E denote B-only/Summary/Execution. $p_H$ uses the relevant 9/9/6 family; $p_{24}$ uses all contrasts.}\label{tab:integratedcontrasts}\\
\toprule Regime & Cmp. & G/L & Gap & Cluster CI & Raw $p$ & $p_H$ & $p_{24}$\\\midrule\endfirsthead
\toprule Regime & Cmp. & G/L & Gap & Cluster CI & Raw $p$ & $p_H$ & $p_{24}$\\\midrule\endhead
\tableinput data/appendix/contrasts.tex
\bottomrule\end{longtable}\endgroup

\paragraph{One execution per cell.}\label{app:noise}
For a fixed regime and target $i$, let $c,c'$ denote two execution conditions, $Y_i(c)$ a binary acceptance result, and $p_i(c)=\Pr[Y_i(c)=1]$ its repeated-run success probability. The quantity $w_i(c)=2p_i(c)(1-p_i(c))$ is the disagreement probability of two independent runs of the same condition. For independent draws $Y_i(c)$ and $Y_i(c')$,
\[
\Pr[Y_i(c)\ne Y_i(c')]=(p_i(c)-p_i(c'))^2+\tfrac12\{w_i(c)+w_i(c')\}.
\]
A single run cannot separate a condition's task-specific effect from execution variation. Resampling quantifies task-composition uncertainty; it does not estimate repeated-run reliability. Representations and shared APEX pairs are not independent replications.

\subsection{Group-level test and recomputed interval sensitivity}
For candidate $c$ and reference $c'$, define the target difference $D_i=Y_i(c)-Y_i(c')$ and group sum $S_g=\sum_{i\in g}D_i$, where $g$ is a domain--family or world group. Let $M$ be the number of groups (15 or 16). We enumerate all $2^M$ sign vectors with components $\epsilon_g\in\{-1,1\}$ and report the fraction satisfying $|\sum_{g=1}^{M}\epsilon_gS_g|\geq|\sum_{g=1}^{M}S_g|$, retaining ties. This sign-flip sensitivity treats groups as independent and requires joint candidate/reference exchangeability within a group, or symmetry of group differences under the null. It is not a causal randomization test. The statistic retains target weighting rather than giving small worlds equal influence. With 15/16 groups, the support is discrete; grouping and multiplicity both reduce precision. No primary contrast survives its global-24 adjustment. Target-level significance and group-level sensitivity should not be conflated.

\path{scripts/analyze_representation_sensitivity.py} computes 20,000 paired and group resamples directly from the frozen cells with seed 20260925 (cohort offsets 0--2, matched-regime offset 3). A group draw retains every target in its sampled groups and divides by the resulting number of targets. All conditions are resampled jointly. The reproducible intervals below are an independently recomputed sensitivity, not the original archived random-number streams used in Figure~\ref{fig:correction_map}. The script also recomputes Execution--Full resource intervals and five matched regime contrasts.

\begingroup\footnotesize\setlength{\tabcolsep}{4pt}
\setlength{\contrastwidth}{\dimexpr\linewidth-12\tabcolsep\relax}
\begin{longtable}{@{}L{.10\contrastwidth}L{.20\contrastwidth}R{.08\contrastwidth}L{.23\contrastwidth}R{.13\contrastwidth}R{.13\contrastwidth}R{.13\contrastwidth}@{}}
\caption{Primary contrasts under group-level inference. CI: freshly recomputed pointwise cluster interval (pp); $p_H$ uses the 9/9/6 families, $p_{24}$ all primary contrasts.}\label{tab:group_sensitivity}\\
\toprule Regime & Hypothesis & Gap & Cluster CI & Group $p$ & $p_H$ & $p_{24}$\\\midrule\endfirsthead
\toprule Regime & Hypothesis & Gap & Cluster CI & Group $p$ & $p_H$ & $p_{24}$\\\midrule\endhead
\tableinput data/representation/primary_cluster.tex
\bottomrule\end{longtable}\endgroup

\subsection{Exploratory representation comparisons}
The 36 within-branch tests cross six unordered representation pairs with U/C and three regimes. They ask which handoff is preferable within a fixed branch. For two formats $r,q\in\{\mathrm{None,Full,Skill,Hybrid}\}$, the separate 18-test family compares $\widehat\Delta_{d,r}^{\mathrm{corr}}-\widehat\Delta_{d,q}^{\mathrm{corr}}$, and therefore answers a different question. No within-branch result survives group-level Holm over 36; three ThinkingBox C-versus-None comparisons survive the target-level version. Only ThinkingBox Full--None survives the group-level 18-test interaction family. These comparisons were motivated by inspecting the existing results and are exploratory.

\begingroup\footnotesize\setlength{\tabcolsep}{4pt}
\setlength{\contrastwidth}{\dimexpr\linewidth-12\tabcolsep\relax}
\begin{longtable}{@{}L{.09\contrastwidth}L{.19\contrastwidth}R{.08\contrastwidth}R{.08\contrastwidth}L{.24\contrastwidth}R{.16\contrastwidth}R{.16\contrastwidth}@{}}
\caption{Within-branch comparisons. N/F/S/H: None/Full/Skill/Hybrid. G/L denotes observed upward/downward transitions. The last columns apply Holm across all 36 tests.}\label{tab:representation_pairs}\\
\toprule Regime & Comparison & G/L & Gap & Cluster CI & Target $p_H$ & Group $p_H$\\\midrule\endfirsthead
\toprule Regime & Comparison & G/L & Gap & Cluster CI & Target $p_H$ & Group $p_H$\\\midrule\endhead
\tableinput data/representation/within_branch.tex
\bottomrule\end{longtable}
\setlength{\LTpre}{0pt}
\setlength{\contrastwidth}{\dimexpr\linewidth-8\tabcolsep\relax}
\begin{longtable}{@{}L{.12\contrastwidth}L{.28\contrastwidth}R{.16\contrastwidth}L{.27\contrastwidth}R{.17\contrastwidth}@{}}
\caption{Difference in correction gaps between representations. All intervals are pp; group $p_H$ adjusts the 18 interactions.}\label{tab:representation_interactions}\\
\toprule Regime & Format contrast & Gap difference & Cluster CI & Group $p_H$\\\midrule\endfirsthead
\toprule Regime & Format contrast & Gap difference & Cluster CI & Group $p_H$\\\midrule\endhead
\tableinput data/representation/interactions.tex
\bottomrule\end{longtable}\endgroup

%% file: source/appendix_rq1.tex
\section{RQ1: Benefit, Harm, Heterogeneity, and Target Errors}\label{app:textresults}\label{app:paths}
\subsection{Complete B-only/U/C outcome decomposition}
With bit order B-only/U/C, 001 is a new success beyond both references, 101 restores a baseline success lost under U, 010 loses U's help, and 110 breaks a shared B-only/U success. Their signed sum is the C--U effect. The other four patterns do not change between U and C. These are observed outcome paths rather than individual causal effects.

\begin{table}[!htb]\centering\small
\caption{All eight outcome patterns, including unchanged outcomes. Each column sums to 100. F/S/H denote Full/Skill/Hybrid.}\label{tab:allpatterns}\label{tab:rq1paths}
\begin{tabular*}{\linewidth}{@{\extracolsep{\fill}}lrrrrrrrrr@{}}\toprule
& \multicolumn{3}{c}{ThinkingBox} & \multicolumn{3}{c}{APEX-state} & \multicolumn{3}{c}{APEX-text}\\
B/U/C & F & S & H & F & S & H & F & S & H\\\midrule
\tableinput data/appendix/triples.tex
\bottomrule\end{tabular*}\end{table}

\paragraph{Conditional improvements and partial credit.}
Let $G=n_{001}+n_{101}$ count gains, $L=n_{010}+n_{110}$ losses, and $Y_i^U=Y_{di}^{U,r}$. The upward proportion among U failures is $G/(N-\sum_iY_i^U)$; the downward proportion among U successes is $L/\sum_iY_i^U$. Full improves 46/77 U failures in ThinkingBox (59.7\%), 8/58 in APEX-state (13.8\%), and 10/55 in APEX-text (18.2\%). It loses 2/23 U successes (8.7\%), 9/42 (21.4\%), and 10/45 (22.2\%), respectively. APEX therefore has substantial headroom, but recovers a smaller fraction of failures while losing a larger fraction of successes.

Aligned rubric differences for Full/Skill/Hybrid are $-4.90/+1.67/+1.29\pp$ in APEX-state and $-2.32/+0.36/-8.26\pp$ in APEX-text. Text Hybrid gains $5.35\pp$ of task-weighted item credit but loses $13.61\pp$. Text Skill improves 49 raw rubric items and loses 34, yet gains only $0.36\pp$ after weighting tasks equally. Counting improved items alone would overstate target-level utility.

\subsection{Relationship and domain heterogeneity}\label{app:relationships}
\paragraph{Request-based relationship coding.}
The 100 frozen pairings shared by both APEX regimes are categorized by reading the complete A and B requests. The mutually exclusive categories use the precedence in Table~\ref{tab:relationshipdefinition}: the same substantive operation takes priority over a shared intermediate procedure, which takes priority over common evidence. Merely using spreadsheets, legal reasoning, or the same application is insufficient to establish a shared procedure. Counts are numbers of pairs, not numbers of distinct families.

\begin{table}[!htb]
\centering\small
\caption{Relationship definitions and examples from the request-based audit.}
\begin{tabularx}{\linewidth}{@{}p{.23\linewidth}rXX@{}}\toprule
Relationship & Pairs & Shared basis & What changes in B\\\midrule
Same-family variant & 42 & Same computation, model, or legal decision test & LBO debt/rate assumptions; a different insured claim\\
Shared intermediate procedure & 26 & Identifiable reusable subprocedure; distinct end operation & Comparable filtering for peer benchmarking versus LBO inputs\\
Shared evidence & 20 & Common case materials or model; distinct operations & Manual drafting versus compliance comparison\\
Context only & 12 & Same project or domain; no specific shared procedure or material established & Board-tenure comparison versus CAPM estimation\\\bottomrule
\end{tabularx}\label{tab:relationshipdefinition}
\end{table}

This is a retrospective, single-pass annotation, not a preregistered or independently blinded classification. Aggregate results were already known; no per-cell target outcomes, gold answers, or rubric scores were used to assign labels. Each pair's shared basis and target change are recorded in \path{data/apex_pair_audit.json}. Borderline abstractions remain judgment-dependent, and no independent agreement estimate is available. The 42 same-family pairs comprise 26 banking, ten legal, and six consulting pairs; relationship and domain are therefore not independently varied.
\begin{table}[!htb]\centering\small
\caption{Matched APEX relationship strata. Every effect is shown as \textbf{state / text} in percentage points. Full/Skill/Hybrid are C--U; E--S is Execution--Summary.}\label{tab:pairedrelations}
\begin{tabular*}{\linewidth}{@{\extracolsep{\fill}}lrrrrr@{}}\toprule
Relationship & Count & Full & Skill & Hybrid & E--S\\\midrule
\tableinput data/regime_pairing/relations_en.tex
\bottomrule\end{tabular*}\end{table}
\begin{table}[!htb]\centering\small
\caption{Domain-specific contrasts, again \textbf{state / text} in percentage points. Same definitions as Table~\ref{tab:pairedrelations}; these are exploratory subgroup results.}\label{tab:paireddomains}
\begin{tabular*}{\linewidth}{@{\extracolsep{\fill}}lrrrrr@{}}\toprule
Domain & Count & Full & Skill & Hybrid & E--S\\\midrule
\tableinput data/regime_pairing/domains_en.tex
\bottomrule\end{tabular*}\end{table}

Let $N_k$ count the pairs in stratum $k$ of a given partition (relationship or domain). The whole-cohort difference is the weighted sum of stratum differences, with weight $N_k/N$ and $N=100$. APEX-state Hybrid's four net successes comprise $-1$ in same-family pairs, $+5$ in shared procedures, $-1$ in shared evidence, and $+1$ in context-only pairs. This locates offsetting changes rather than identifying a causal effect of relationship type. Shared procedures span 12 worlds (11 banking, two law, and 13 consulting pairs); shared evidence spans 11 worlds (six banking, eight law, and six consulting pairs).

\begin{figure}[!htb]\centering
\resizebox{\linewidth}{!}{\input{figures/rq1_heterogeneity_en}}
\caption{APEX C--U differences (pp); parentheses: subgroup sizes. Domains and relationships separately partition the same 100 pairs. Color describes differences, not significance.}\label{fig:heterogeneity_main}
\end{figure}
Among the 42 same-family pairs, B-only solves twenty in each regime. State C-Full/Skill/Hybrid solve 14/17/16, versus 21/19/18 for text. Text Full's seven additional successes are not a correction gain: its U and C both solve 21. In shared procedures, state Hybrid and text Skill each gain five successes, whereas text Hybrid loses one. Relatedness provides an opportunity for reuse, not a guarantee that a source choice applies.

Legal Skill moves from eight to twelve successes with state and from four to eleven with text. The larger text correction gap reflects a weaker U, not a stronger C. Banking Hybrid moves from twenty to twenty with state and from 26 to seventeen with text, with no gains and nine losses. This combines six additional U successes with three fewer C successes; it is not nine C failures attributable to removing state. The text banking contrast has raw $p=0.003906$ and twelve-test domain-family $p_H=0.046875$; subgroup analyses remain exploratory.

\subsection{Consistency across representations and handoff regimes}
\begin{table}[!htb]\centering\small
\caption{Targets improved or regressed under at least one, at least two, or all three C--U representations. Improvement and regression columns are not disjoint: a target can improve under one representation and regress under another.}\label{tab:consistency}
\begin{tabular*}{\linewidth}{@{\extracolsep{\fill}}lrrrrrr@{}}\toprule
& \multicolumn{3}{c}{Improved} & \multicolumn{3}{c}{Regressed}\\
Regime & $\geq1$ & $\geq2$ & All 3 & $\geq1$ & $\geq2$ & All 3\\\midrule
\tableinput data/appendix/consistency.tex
\bottomrule\end{tabular*}\end{table}
\begin{table}[!htb]\centering\small
\caption{Matching the \emph{same target's contrast} across APEX regimes. Gains/losses mean C--U, except E--S. ``G to L'' means a state gain becomes a text loss; ``L to G'' is the reverse. Each row has 100 matched pairs.}\label{tab:gainoverlap}
\begin{tabular*}{\linewidth}{@{\extracolsep{\fill}}lrrrrrr@{}}\toprule
Contrast & State gains & Text gains & Gain in both & Loss in both & G to L & L to G\\\midrule
\tableinput data/regime_pairing/overlap_rows.tex
\bottomrule\end{tabular*}\end{table}
\begin{table}[!htb]\centering\small
\caption{Within-regime effects and matched differences of effects. The last column is a world-cluster 95\% interval for text minus state, in percentage points.}\label{tab:regimedifference}
\begin{tabular*}{\linewidth}{@{\extracolsep{\fill}}lrrrl@{}}\toprule
Contrast & State & Text & Difference & World interval\\\midrule
C-Full--U-Full & -1 & 0 & +1 & $[-6.19,6.67]$\\
C-Skill--U-Skill & +3 & +1 & -2 & $[-15.74,12.77]$\\
C-Hybrid--U-Hybrid & +4 & -8 & -12 & $[-24.00,1.10]$\\
Execution--Summary & +4 & +12 & +8 & $[-5.46,24.27]$\\
Full--Execution & -3 & -7 & -4 & $[-19.80,7.56]$\\\bottomrule
\end{tabular*}\end{table}

APEX B-only passes in both regimes for 37 targets, in state only for seven, and in text only for ten; 46 fail in both. Even without any A-derived handoff, observed outcomes change across executions. The matched contrasts compare correction-associated changes over and above each regime's own baseline, rather than treating a difference between two condition means as a state effect.

The matched gap difference $\widehat\delta_r=\widehat\Delta_{\mathrm{text},r}^{\mathrm{corr}}-\widehat\Delta_{\mathrm{state},r}^{\mathrm{corr}}$ compares correction effects on the same 100 pairs. It is $+1/-2/-12\pp$ for Full/Skill/Hybrid, with every world interval spanning zero (Table~\ref{tab:regimedifference}). It compares gaps, not C endpoints: separate significance does not establish an interaction. Text construction and execution batches also differ, so this comparison does not isolate state inheritance.

Similar totals need not identify the same beneficiaries. Only two of state Full's eight gains overlap with text Full's ten; Skill shares five gains and Hybrid two. Hybrid has four targets changing from a state gain to a text loss, with none in the reverse direction. Full has no direct gain-to-loss reversal: its differences involve gains or losses becoming ties, or vice versa (Table~\ref{tab:gainoverlap}).

\subsection{Cohort-wide validator profiles and cluster influence}\label{app:validator}
We export the saved validator paths for all 1,100 ThinkingBox cells, including successes, from the same files used for acceptance. A lifecycle flag marks a path beginning with \texttt{zendesk\_ticket\_status\_violations}; a type flag marks \texttt{zendesk\_tickets[k].type}. All failed cells have recorded difference paths. These two diagnostics are overlapping subsets of benchmark errors, not an exhaustive failure taxonomy. The flags refer to the target validator; they do not label whether the source feedback explicitly taught that requirement.

\begin{table}[!htb]
\centering\small
\caption{Recorded lifecycle violations, all 100 paired targets in each comparison. ``Removed'' counts reference-flagged targets without the flag in the candidate. Its last two columns partition those targets by candidate acceptance.}
\begin{tabular*}{\linewidth}{@{\extracolsep{\fill}}lrrrrr@{}}\toprule
Comparison & Before & After & Removed & Then pass & Still fail\\\midrule
\tableinput data/validator_transition_rows.tex
\bottomrule\end{tabular*}\label{tab:validator}
\end{table}
New flags also appear: one for C-Full, six for C-Skill, three for C-Hybrid, and one for Execution, relative to their respective references. Thus removing flags on some targets is compatible with introducing them elsewhere. The identity ``removed = then pass + still fail'' distinguishes correcting a recorded local error from satisfying every task requirement. It does not estimate a causal mediation effect. \path{data/validator_profiles.json} retains every pair's paths; \path{data/validator_analysis.json} includes both flag types and the paired assignments.

\begin{table}[!htb]\centering\small
\caption{Leave-one-cluster-out acceptance gaps (pp), including all three regimes. Each row removes one complete family or world at a time. Min/max are influence ranges, not confidence intervals. Abbreviations follow Table~\ref{tab:integratedcontrasts}.}\label{tab:clusterinfluence}
\begin{tabular*}{\linewidth}{@{\extracolsep{\fill}}llrrr@{}}\toprule
Regime & Comparison & Clusters & Minimum & Maximum\\\midrule
\tableinput data/appendix/influence.tex
\bottomrule\end{tabular*}\end{table}
ThinkingBox's C--U advantage remains positive after omitting any one family. Both APEX regimes retain substantially smaller effects. Text Execution--Summary remains positive after omitting any one world; this influence result is distinct from multiplicity-adjusted testing.

%% file: figures/rq1_heterogeneity_en.tex
\begin{tikzpicture}[x=1cm,y=0.8cm,font=\sffamily\fontsize{8}{9}\selectfont]
\definecolor{ink}{HTML}{243447}
\definecolor{newgain}{HTML}{188D89}
\definecolor{restore}{HTML}{80BDDC}
\definecolor{losthelp}{HTML}{EBA88C}
\definecolor{broken}{HTML}{C35562}
\node[font=\sffamily\bfseries\small] at (6.050,4.650) {APEX-state};
\node[] at (4.650,4.250) {Full};
\node[] at (6.050,4.250) {Skill};
\node[] at (7.450,4.250) {Hybrid};
\node[anchor=east] at (3.700,3.760) {Banking (48)};
\fill[broken!37.5!white] (4.000,3.560) rectangle (5.320,3.960);
\node[] at (4.660,3.760) {-10.4};
\fill[newgain!0.0!white] (5.400,3.560) rectangle (6.720,3.960);
\node[] at (6.060,3.760) {+0.0};
\fill[newgain!0.0!white] (6.800,3.560) rectangle (8.120,3.960);
\node[] at (7.460,3.760) {+0.0};
\node[anchor=east] at (3.700,3.300) {Law (23)};
\fill[newgain!15.7!white] (4.000,3.100) rectangle (5.320,3.500);
\node[] at (4.660,3.300) {+4.3};
\fill[newgain!62.6!white] (5.400,3.100) rectangle (6.720,3.500);
\node[] at (6.060,3.300) {+17.4};
\fill[newgain!62.6!white] (6.800,3.100) rectangle (8.120,3.500);
\node[] at (7.460,3.300) {+17.4};
\node[anchor=east] at (3.700,2.840) {Consulting (29)};
\fill[newgain!37.2!white] (4.000,2.640) rectangle (5.320,3.040);
\node[] at (4.660,2.840) {+10.3};
\fill[broken!12.4!white] (5.400,2.640) rectangle (6.720,3.040);
\node[] at (6.060,2.840) {-3.4};
\fill[newgain!0.0!white] (6.800,2.640) rectangle (8.120,3.040);
\node[] at (7.460,2.840) {+0.0};
\node[anchor=east] at (3.700,2.230) {Same-family (42)};
\fill[broken!17.1!white] (4.000,2.030) rectangle (5.320,2.430);
\node[] at (4.660,2.230) {-4.8};
\fill[newgain!25.7!white] (5.400,2.030) rectangle (6.720,2.430);
\node[] at (6.060,2.230) {+7.1};
\fill[broken!8.6!white] (6.800,2.030) rectangle (8.120,2.430);
\node[] at (7.460,2.230) {-2.4};
\node[anchor=east] at (3.700,1.770) {Shared procedure (26)};
\fill[newgain!0.0!white] (4.000,1.570) rectangle (5.320,1.970);
\node[] at (4.660,1.770) {+0.0};
\fill[newgain!0.0!white] (5.400,1.570) rectangle (6.720,1.970);
\node[] at (6.060,1.770) {+0.0};
\fill[newgain!69.2!white] (6.800,1.570) rectangle (8.120,1.970);
\node[] at (7.460,1.770) {+19.2};
\node[anchor=east] at (3.700,1.310) {Shared evidence (20)};
\fill[broken!18.0!white] (4.000,1.110) rectangle (5.320,1.510);
\node[] at (4.660,1.310) {-5.0};
\fill[broken!18.0!white] (5.400,1.110) rectangle (6.720,1.510);
\node[] at (6.060,1.310) {-5.0};
\fill[broken!18.0!white] (6.800,1.110) rectangle (8.120,1.510);
\node[] at (7.460,1.310) {-5.0};
\node[anchor=east] at (3.700,0.850) {Context only (12)};
\fill[newgain!60.0!white] (4.000,0.650) rectangle (5.320,1.050);
\node[] at (4.660,0.850) {+16.7};
\fill[newgain!30.0!white] (5.400,0.650) rectangle (6.720,1.050);
\node[] at (6.060,0.850) {+8.3};
\fill[newgain!30.0!white] (6.800,0.650) rectangle (8.120,1.050);
\node[] at (7.460,0.850) {+8.3};
\draw[gray!70] (4.000,2.540) -- (8.120,2.540);
\node[font=\sffamily\bfseries\small] at (11.250,4.650) {APEX-text};
\node[] at (9.850,4.250) {Full};
\node[] at (11.250,4.250) {Skill};
\node[] at (12.650,4.250) {Hybrid};
\fill[broken!22.5!white] (9.200,3.560) rectangle (10.520,3.960);
\node[] at (9.860,3.760) {-6.2};
\fill[broken!22.5!white] (10.600,3.560) rectangle (11.920,3.960);
\node[] at (11.260,3.760) {-6.2};
\fill[broken!67.5!white] (12.000,3.560) rectangle (13.320,3.960);
\node[] at (12.660,3.760) {-18.8};
\fill[newgain!0.0!white] (9.200,3.100) rectangle (10.520,3.500);
\node[] at (9.860,3.300) {+0.0};
\fill[newgain!78.0!white] (10.600,3.100) rectangle (11.920,3.500);
\node[] at (11.260,3.300) {+30.4};
\fill[newgain!15.7!white] (12.000,3.100) rectangle (13.320,3.500);
\node[] at (12.660,3.300) {+4.3};
\fill[newgain!37.2!white] (9.200,2.640) rectangle (10.520,3.040);
\node[] at (9.860,2.840) {+10.3};
\fill[broken!37.2!white] (10.600,2.640) rectangle (11.920,3.040);
\node[] at (11.260,2.840) {-10.3};
\fill[newgain!0.0!white] (12.000,2.640) rectangle (13.320,3.040);
\node[] at (12.660,2.840) {+0.0};
\fill[newgain!0.0!white] (9.200,2.030) rectangle (10.520,2.430);
\node[] at (9.860,2.230) {+0.0};
\fill[broken!25.7!white] (10.600,2.030) rectangle (11.920,2.430);
\node[] at (11.260,2.230) {-7.1};
\fill[broken!34.3!white] (12.000,2.030) rectangle (13.320,2.430);
\node[] at (12.660,2.230) {-9.5};
\fill[newgain!0.0!white] (9.200,1.570) rectangle (10.520,1.970);
\node[] at (9.860,1.770) {+0.0};
\fill[newgain!69.2!white] (10.600,1.570) rectangle (11.920,1.970);
\node[] at (11.260,1.770) {+19.2};
\fill[broken!13.8!white] (12.000,1.570) rectangle (13.320,1.970);
\node[] at (12.660,1.770) {-3.8};
\fill[newgain!18.0!white] (9.200,1.110) rectangle (10.520,1.510);
\node[] at (9.860,1.310) {+5.0};
\fill[broken!18.0!white] (10.600,1.110) rectangle (11.920,1.510);
\node[] at (11.260,1.310) {-5.0};
\fill[broken!18.0!white] (12.000,1.110) rectangle (13.320,1.510);
\node[] at (12.660,1.310) {-5.0};
\fill[broken!30.0!white] (9.200,0.650) rectangle (10.520,1.050);
\node[] at (9.860,0.850) {-8.3};
\fill[newgain!0.0!white] (10.600,0.650) rectangle (11.920,1.050);
\node[] at (11.260,0.850) {+0.0};
\fill[broken!60.0!white] (12.000,0.650) rectangle (13.320,1.050);
\node[] at (12.660,0.850) {-16.7};
\draw[gray!70] (9.200,2.540) -- (13.320,2.540);
\node[font=\sffamily\scriptsize] at (7.000,-0.070) {C minus U (pp)   |   red: lower acceptance     teal: higher acceptance};
\end{tikzpicture}

%% file: source/appendix_cases.tex
\section{Action-Level Evidence for Transfer and Its Limits}\label{app:cases}\label{app:extendedmechanisms}
\paragraph{ThinkingBox: a reusable constraint can outlive the source decision.}
U-Full records 55 ticket-lifecycle violations; C-Full records two. Of 54 reference-flagged targets without that flag in C, 37 pass and 17 still fail; one new flag appears elsewhere. This distinguishes improvement in an operational requirement from full target acceptance. In \texttt{lif\_017}$\to$\texttt{lif\_015}, the business eligibility decision changes, but the open--act--close requirement remains. B-only and U-Skill correctly decline B while creating already-solved tickets; C-Skill declines it with the required lifecycle. The source concerns an unpaid Preferred-tier customer eight days after cancellation, within a 30-day reinstatement window; B concerns a paid customer 35 days after cancellation. All inspected B-only/U-Skill/C-Skill runs correctly decline B, separating the changed business decision from the reusable ticket workflow. By contrast, a source correction from Daughter to Child in \texttt{drv\_101}$\to$\texttt{drv\_008} cannot be claimed as field-level transfer when B rejects the request and never writes that field. Episode usefulness and transfer of a specific correction are different claims.

\paragraph{APEX: a source-convention conflict occurs under both handoffs.}
For \texttt{ad52b8019140}, B-only and all three U representations pass in \emph{both} regimes, while all three C representations fail with zero rubric credit. B explicitly specifies mid-year convention. In the text trace, C-Skill labels full-year terminal-value discounting as user-validated. B reads the original model, calculates both six-year and 5.5-year alternatives, then selects the source-validated six-year result: 40.45/44.83 instead of B's fixed-rubric 41.62/46.22. The observed conflict concerns applicability and selection, not missing retrieval or calculation. The same score pattern occurs without inherited A files; the detailed selection chain is established from the text run. It does not isolate the approval marker's causal contribution.

\paragraph{Positive transfer can mix a method with a source conclusion.}
For \texttt{327f2507ef39}, B-only and all U variants fail in both regimes. State C-Full and C-Skill pass while C-Hybrid fails; all three text C variants pass. C-Skill prescribes a requirements--source-clause matrix, entity-role separation, and avoiding defects imported from unrelated documents. It also retains A's conclusion that the reviewed documents comply. B uses the matrix and reaches the rubric's compliant conclusion. This is useful corrected experience in both regimes, with representation-sensitive coverage; it does not separate procedural learning from answer anchoring.

\paragraph{Restoration and loss of U's help are also visible in both regimes.}
Helios (\texttt{993a38dad720}) has B-only success and a U-Hybrid failure followed by C-Hybrid success in both state and text. The text artifact audit identifies the changed data scope: U unions 400-, 2,000-, and 2,500-row inputs into 4,900 SKUs and reports a 22.06\% VW share; C uses the specified 2,500-row Rebuilt input, with 541 VW entries and 21.64\%. This restores an independently achievable result by applying an appropriate scope constraint.

Lumea (\texttt{e8b80949477a}) shows the opposite result in both regimes: U-Skill and U-Hybrid pass, both corresponding C conditions receive 1/3, and B-only also receives 1/3. The text trace computes alternatives from workbook precision and rounded CSV values but selects the latter. Weighted margins are 56.786890\% versus 56.795421\%; the 30\% improvement opportunity is 607,500.34 versus 606,666.14. A shared displayed margin of 56.8\% hides a consequential precision choice. This is a reference-sensitive loss relative to U, not a loss of observed baseline success; the score alone does not establish harmful reasoning.

These cases expose a common decision point: \textbf{which source constraint should govern the target?} Correct information can be available but unenacted; an enacted rule can be inappropriate; a corrected local behavior can still leave other requirements unsatisfied. The observed ThinkingBox benefit is compatible with reusable workflow constraints, whereas APEX cases expose version, scope, precision, and convention selection. We do not infer a benchmark-wide causal explanation from those cases.

\paragraph{State availability versus correction benefit.}
Physical state changes what must be reconstructed, but does not decide which result is applicable. In Planet Fitness (\texttt{01da0d93b3ad}), the source text names v08 while the text target has v06; Summary and Execution check availability and must rebuild missing work. In the core conditions, state U/C-Full both fail, while text U/C-Full both pass. This level change creates no C--U improvement in either regime. Thus successful reconstruction is not automatically evidence of correction transfer.
\paragraph{ThinkingBox disagreement set.}
All 15 Execution gains against Summary and all four losses are included in \path{data/case_evidence.json}, with task IDs and validator paths from the failed execution. The eight gains with ticket-status violations are \texttt{veh\_013}, \texttt{bil\_016}, \texttt{drv\_021}, \texttt{drv\_101}, \texttt{veh\_005}, \texttt{bil\_014}, \texttt{exp\_011}, and \texttt{hwa\_013}. Ticket-type mismatches occur in \texttt{ldr\_008} and \texttt{bil\_010}. The other gains are \texttt{bil\_008}, \texttt{trv\_009}, \texttt{csa\_004}, \texttt{exp\_002}, and \texttt{onb\_007}. Secondary timestamp differences are not counted as separate explanatory mechanisms. The four losses are \texttt{lif\_012} (incorrect billing-state change), \texttt{swa\_013} and \texttt{onb\_011} (missing approval fields), and \texttt{doc\_001} (ticket-status violation). The coding is based on outcome-selected validator differences, not blinded causal annotation.

\paragraph{Insurance rejection: business decision versus operational contract.}
The \texttt{ldr\_008} target belongs to \texttt{sandbox\_auto\_insurance\_group1.py}. In both executions the agent rejects a billing change for a listed driver lacking policyholder authority. The supplied source summary explicitly states ``type task''. Nevertheless, the Summary-condition target updates the ticket to a solved rejection without setting its type; Execution sets type to task and completes its status update. The other ticket-type gain, \texttt{bil\_010}, also receives a summary explicitly specifying ``type task''. Both errors occur despite that field value being available, rather than because compression removed it. \path{data/case_evidence.json} includes the supplied summary and source task for all 19 disagreements. These comparisons establish an availability--enactment distinction; they do not isolate whether repetition, action formatting, or another feature of the added trace drives it.

\paragraph{APEX-state valuation: a common upstream source affects four criteria.}
Pair \texttt{apex\_full\_3724da87dc96} requests a downside deck for 3M's 20\% stake using a 10\% free-cash-flow reduction. The following values are taken from delivered answers and rubric evidence.
\begin{table}[!htb]
\centering\small
\caption{3M downside analysis. Dollar values are millions. All conditions produce a deck; model lineage changes the numerical findings.}
\begin{tabularx}{\linewidth}{@{}Xrrr@{}}\toprule
Requested quantity & Summary & Execution & Reference\\\midrule
Current stake & 4,935.9 & 5,499.7 & 5,499.7\\
PV of revised cash flows & 4,389.4 & 4,790.1 & 4,790.1\\
Discounted terminal value & 17,822.9 & 19,959.1 & 19,959.0\\
Revised stake & 4,442.5 & 4,949.8 & 4,949.8\\
Percentage loss & 10.0\% & 10.0\% & 10.0\%\\\midrule
Criteria satisfied & 1/5 & 5/5 & --\\\bottomrule
\end{tabularx}
\end{table}
Summary uses the \texttt{v1 - FIXED} workbook, whereas Execution's primary source is the original \texttt{v1.xlsx}, \texttt{DCF-Solv} tab. Execution also computes the corrected copy as a labeled cross-check. C-Full yields 1/5 and a current stake of approximately \$4,936.4M using the corrected lineage. B requests values ``directly from the accretion dilution model,'' a 20\% ownership stake, and a 10\% reduction in 2025--2029 cash flows, but does not name the original or corrected version. The saved reads identify a lineage-dependent score difference; they do not establish which interpretation a user would intend after A's accepted repair. The fixed-reference outcome is retained without treating it as proof that correction history harms reasoning.

\paragraph{Distinguishing inherited errors from new target errors.}
In a separate ThinkingBox equipment case, \texttt{hwa\_013}, the C-Skill run assigns its second reserved device rather than its first. Neither identifier appears in the source handoff: both originate in actual B-stage tool returns. The first reservation remains orphaned and a ticket-status violation is recorded. This is a new execution-side divergence, not evidence that memory copied a stale identifier. The packaged supplementary evidence records this provenance.

\subsection{Cross-family evidence and ambiguous score differences}\label{app:crossfamily}
Shared-procedure APEX-state Hybrid has six C--U gains and one loss (raw $p=0.125$), compared with four gains and one loss against B-only ($p=0.375$). In peer-screening pair \texttt{4e38ae1e886d}, both branches exclude the same comparables and calculate the same 20.67x median, 30.67x exit multiple, and 3.42x debt multiple. The difference arises later in LBO returns, so it does not demonstrate learning the common screening step. In \texttt{432cf30bbe45}, review explanations invoke different acquisition-rate targets (4.9\% versus 4.1\% after tax); omitting that pair leaves five gains and one loss among 25 pairs ($+16.0\pp$, raw $p=0.219$).

All three shared-evidence Execution--Summary gains were inspected. In Kenvue (\texttt{6fa5694c8bbe}), Execution selects volume-weighted closing prices while Summary selects arithmetic means, changing rubric satisfaction from 3/5 to 5/5. Summary also computes weighted alternatives, and neither source handoff teaches VWAP. In Aptar (\texttt{f58e573d229d}), both answers headline 36.40\% and provide 28.61 points under an alternative interpretation. In purchase-price allocation (\texttt{32223647d439}), Execution's review marks goodwill passed while its evidence describes a mismatch. The latter two are not behaviorally corroborated gains; omitting both leaves one gain and no loss among 18 pairs ($+5.6\pp$, raw $p=1$). Original outcomes remain intact.

These selected audits distinguish transferred procedure, source-answer anchoring, target-side evidence selection, and judge disagreement. They are evidence about particular action chains, not estimates of the prevalence of each mechanism.

%% file: source/appendix_retention.tex
\section{RQ2 and RQ3: Accepted Execution and Resource Tradeoffs}\label{app:retention}
\begin{table}[!htb]\centering\small
\caption{Accepted-experience conditions. G/L counts paired gains/losses for Execution--Summary; $p_H$ is the six-test H2-family adjustment.}\label{tab:regimeboundary}
\begin{tabular*}{\linewidth}{@{\extracolsep{\fill}}lrrrrrl@{}}\toprule
Regime & Summary & Execution & Full & G/L & E--S & $p_H$\\\midrule
ThinkingBox & 56 & 67 & 67 & 15/4 & +11 & 0.0961\\
APEX-state & 40 & 44 & 41 & 12/8 & +4 & 1.0000\\
APEX-text & 40 & 52 & 45 & 14/2 & +12 & 0.0251\\\bottomrule
\end{tabular*}\end{table}

\begin{table}[!htb]\centering\small
\caption{B-stage resources for B-only and corrected handoffs. Each mean includes all 100 targets, including failures. Input is cumulative millions of tokens; input/call is total input divided by total calls in thousands.}\label{tab:resourcealternatives}\label{tab:regimecost}
\begin{tabular*}{\linewidth}{@{\extracolsep{\fill}}llrrrrr@{}}\toprule
Regime & Condition & Accepted & Input M & Calls & Input/call k & Tools\\\midrule
\tableinput data/appendix/resources.tex
\bottomrule\end{tabular*}\end{table}
\subsection{Resource thresholds and initial payloads}

Execution reduces mean cumulative input relative to Full by 18.7\% in ThinkingBox, 12.1\% in APEX-state, and 31.8\% in APEX-text. Mean model calls instead increase by 4.7\%, 105.7\%, and 16.9\%. H3a is consistent across observed means; H3b has the opposite direction throughout. The text cost analysis gives a world interval of $[-1.607,-0.728]$ million input tokens and $[1.10,7.94]$ additional calls, with two-test H3 Holm values 0.0002 and 0.0068.

The accounting is multiplicative: cumulative input equals calls times average input per call. In state APEX, Execution roughly halves input per call while doubling calls, leaving only a 12.1\% net input reduction. Text APEX reduces input per call from 119.2k to 69.5k, enough to offset its smaller call increase. Tool calls in text rise from 38.24 to 44.40, while output tokens remain approximately 56k. A shorter handoff changes how much context is repeatedly read and how much work remains downstream; it does not uniformly shorten execution.

The cost interpretation also depends on the reference. Text Execution uses 12.2\% more input than Summary (2.438M versus 2.172M), while its 2.21-call reduction has a world interval crossing zero. Skill and Hybrid reduce input relative to Full but require more calls and do not dominate its acceptance. At an observed completed-trajectory ceiling of two million input tokens, text Execution has 32 accepted targets versus Full's 11; at twenty model calls, Full has 18 versus Execution's ten. These are counts of completed accepted runs within resource bounds, not early-stopping experiments.

Input includes cached context, so token reductions are not billing reductions. Initial payload and whole initial-message length are also distinct: the text source handoff averages 281.2k characters for Full versus 45.3k for Execution, while serialized initial messages average 289.7k and 48.8k. Costs cover the current B attempts, excluding source correction, memory construction, judging, and historical infrastructure retries.

%% file: source/appendix_judging.tex
\section{Judging, Coverage, and Sensitivity Analyses}\label{app:coverage}\label{app:textalignment}
\subsection{Full-denominator coverage and artifact reconstruction}
\begin{table}[!htb]\centering\small
\caption{APEX execution coverage. Completed means agent submission, not owner acceptance. Budget and infrastructure failures remain distinct. Every cell has a judgment in the aligned analysis.}
\begin{tabular*}{\linewidth}{@{\extracolsep{\fill}}lrrrr@{}}\toprule
Regime / set & Completed & Budget failed & Error & Total\\\midrule
State: original nine conditions & 895 & 5 & 0 & 900\\
State: Summary & 92 & 8 & 0 & 100\\
State: Execution & 88 & 10 & 2 & 100\\
Text: all eleven conditions & 1,072 & 28 & 0 & 1,100\\\bottomrule
\end{tabular*}\end{table}
State judgments reconstruct the original world, A inputs and branch delta, B inputs, and B delta. Text judgments reconstruct the original world, B inputs, and the text run's B delta without restoring A. Both apply the original target rubric to native visual evidence and available artifacts. Technical rendering, transport, or response-format errors are not scored as task failures. Valid judgments are retained rather than repeated because of a low score.

\subsection{Partial-artifact alignment and execution-error sensitivity}
The original state archive contains 1,057 primary \texttt{deepseek-flash} reviews, 23 \texttt{gpt-5.6-luna} fallback reviews, and 20 policy-zero cells. The latter comprise 18 budget failures and two execution errors. An independent alignment layer reviews their saved partial artifacts with \texttt{deepseek-flash}; all 20 remain unaccepted. Summary's mean rubric score changes from 49.22\% to 49.65\%; Execution remains 52.33\%. The current condition table uses these aligned rubric values. Original reviews, policy-zero scores, and execution identities are preserved.

For H2 contrasts involving Execution, excluding the two state execution-error pairs leaves 98 matched pairs. Execution--Summary is $+5.10\pp$ in state and $+12.24\pp$ in text, a matched regime difference of $+7.14\pp$, compared with $+8\pp$ over all 100. Full--Execution is $-4.08\pp$ and $-7.14\pp$, respectively, a difference of $-3.06\pp$. These exclusions do not convert the regime comparison into a randomized state intervention.

\subsection{Backend and disagreement-audit sensitivity}
APEX-text has 1,083 valid primary reviews and 17 \texttt{gpt-6-luna} fallbacks for technical failures. Sixteen fallback cells retain verified native-image tool traces; the first valid fallback has incomplete image-access audit provenance. Among the 98 pairs with primary-backend reviews for both Summary and Execution, the gap is $+12.24\pp$ (14 gains, two losses). This subset is not evidence that judge backends are interchangeable.

The complete 16-case APEX-text Execution--Summary disagreement audit separates requirement coverage, formal delivery, answer priority, numerical precision, and scoring contradictions. Four clearly inconsistent judgments are retained as a separate sensitivity layer. Applying those audit corrections gives 50/38 rather than 52/40 accepted targets, leaving a twelve-point gap but changing 14/2 discordances to 12/0. Main results retain the first valid judgments. Because this audit selects disagreements and is not condition-blinded, it does not estimate overall judge accuracy or provide a second independent test.

Text Execution--Summary also improves rubric satisfaction by $8.96\pp$; its adjusted $p=0.0568$ is from the separate eight-contrast text rubric family, not the acceptance families in Table~\ref{tab:integratedcontrasts}. Acceptance, partial credit, backend restriction, and action-level audit answer different questions and are reported separately.

%% file: source/appendix_related.tex
\section{Extended Literature Comparison}
\label{app:related}
\paragraph{Correction in continuing human--AI collaboration.}
\citet{amershi2019guidelines} distinguish making errors easy to correct, remembering recent interaction, and adapting from user behavior over time. Our question concerns their intersection: does an accepted repair improve a new case, or must its lesson be supplied again? We operationalize this through paired agent executions with simulated feedback, rather than a human-subject evaluation of interface usability or oversight effort.

\paragraph{Improving an attempt versus improving the next task.}
Self-Refine \citep{madaan2023selfrefine} improves outputs through self-feedback; Reflexion \citep{shinn2023reflexion} turns task feedback into reflections for later attempts. Self-correction without external feedback need not improve reasoning \citep{huang2024selfcorrect}, and early mistaken assumptions can persist across turns \citep{laban2025lost}. We examine what happens \emph{after} A is accepted: how corrected and uncorrected experience affect a distinct B.

\paragraph{Evidence for cross-task experience reuse.}
ExpeL \citep{zhao2024expel} extracts knowledge from training-task experience for inference-time retrieval. CLIN \citep{majumder2024clin} updates causal abstractions and cross-episode meta-memory for new tasks and environments. Agent Workflow Memory (AWM) \citep{wang2025awm} induces workflows for offline, online, and cross-domain use on Mind2Web and WebArena. ReasoningBank \citep{ouyang2026reasoningbank} distills self-judged successes and failures; memory-aware test-time scaling (MaTTS) expands exploration and ablates failed-trajectory contributions. We do not propose a larger skill library or a new memory learner. We pair one failed episode with its accepted repair and hold B fixed, asking what changes downstream when the source is corrected. Summary and execution controls then locate how the corrected experience is conveyed.

\paragraph{Skills can also hurt.}
Trace2Skill \citep{ni2026trace2skill} aggregates trajectory-local lessons and tests cross-model and out-of-distribution transfer. \emph{Break It Down, Pass It On} \citep{feng2026transfer} compares induction granularity and text/code formats: task-level skills often hurt, while subtask-level skills help on average, with specificity and abstractness characterizing utility. Its analyses also examine source outcomes and intervene on skill-library utility; applicability and negative transfer are not unique to our study. SkillsBench \citep{li2026skillsbench} pairs no-skill and curated-skill conditions, also tests self-generated skills, and analyzes gains and regressions. Our added axis pairs the \emph{same episode before and after accepted repair}, separating an update's incremental value from the corrected memory's value relative to independence (Table~\ref{tab:prior}).

\paragraph{Compression and inference-time adaptation.}
LLMLingua \citep{jiang2023llmlingua} compresses prompts; Lost in the Middle \citep{liu2024middle} studies positional effects in long-context use. The Complexity Trap \citep{lindenbauer2025complexity} compares observation masking with model summaries, while Evaluating AGENTS.md \citep{gloaguen2026agents} measures benefits and costs of repository context files. Trace as State \citep{zou2026trace} prepends reasoning traces to support rereading within a task. We fix model parameters and study experience utility and execution cost across a task boundary.

\paragraph{Evaluation granularity and reliability.}
MemoryArena \citep{he2026memoryarena} evaluates memory-guided action across interdependent sessions; we additionally vary source correction. Micro-benchmarking examines ranking reliability under small gaps \citep{yauney2026micro}. We decompose outcomes through acceptance, rubric satisfaction, and paired trajectories.

\begin{table}[!htb]
\centering\footnotesize
\caption{Comparison units and experimental axes in cross-task experience reuse.}
\begin{tabularx}{\linewidth}{@{}lXX@{}}\toprule
Study & How experience transfers & Main comparison axis\\\midrule
ExpeL \citep{zhao2024expel} & Knowledge extraction and experience retrieval & Accumulation and transfer\\
CLIN \citep{majumder2024clin} & Causal abstractions and meta-memory & Adaptation and generalization\\
AWM \citep{wang2025awm} & Induced reusable workflows & Offline/online and cross-domain\\
ReasoningBank \citep{ouyang2026reasoningbank} & Strategies from success and failure & Memory; exploration via MaTTS\\
\citet{feng2026transfer} & Task- or subtask-induced skills & Granularity $\times$ text/code\\
SkillsBench \citep{li2026skillsbench} & Curated or generated skills & Skill source, domain, and utility\\
This study & The same source failure with and without an accepted repair, transferred to a fixed B & Memory-update value versus predecessor and independence; repair $\times$ representation\\\bottomrule
\end{tabularx}\label{tab:prior}
\end{table}